\documentclass[twocolumn]{svjour3}
\usepackage{amsmath,amsfonts,amssymb}
\usepackage{ijcv-layout}
\usepackage{algorithmic}
\usepackage{algorithm}
\usepackage{textcomp}
\usepackage{verbatim}
\usepackage[mathscr]{euscript}
\usepackage{cleveref}
\crefname{figure}{Fig.}{Figs.}
\crefname{table}{Table}{Tables}
\crefname{section}{Sect.}{Sects.}
\crefname{equation}{Eq.}{Eqs.}
\renewcommand{\cite}{\unskip\space\citep}

\newif\ifrevision
\revisionfalse

\newcommand{\rev}[1]{%
    \ifrevision
        \textcolor{red}{#1}%
    \else
        #1%
    \fi
}

\makeatletter

\let\revrow@font\relax

\ifrevision
\patchcmd{\insert@column}
    {\@sharp}
    {\revrow@font\@sharp}
    {}
    {\PackageError{revrow}{Failed to patch tabular}{}}

\patchcmd{\insert@pcolumn}
    {\@sharp}
    {\revrow@font\@sharp}
    {}
    {}

\CT@everycr\expandafter{%
    \the\CT@everycr
    \noalign{\global\let\revrow@font\relax}%
}
\fi

\newcommand{\revrow}{%
    \ifrevision
        \global\def\revrow@font{\color{red}}%
        \color{red}
    \fi
}

\makeatother

\newcommand{\insertfig}
{

\includegraphics[width=0.98\linewidth]{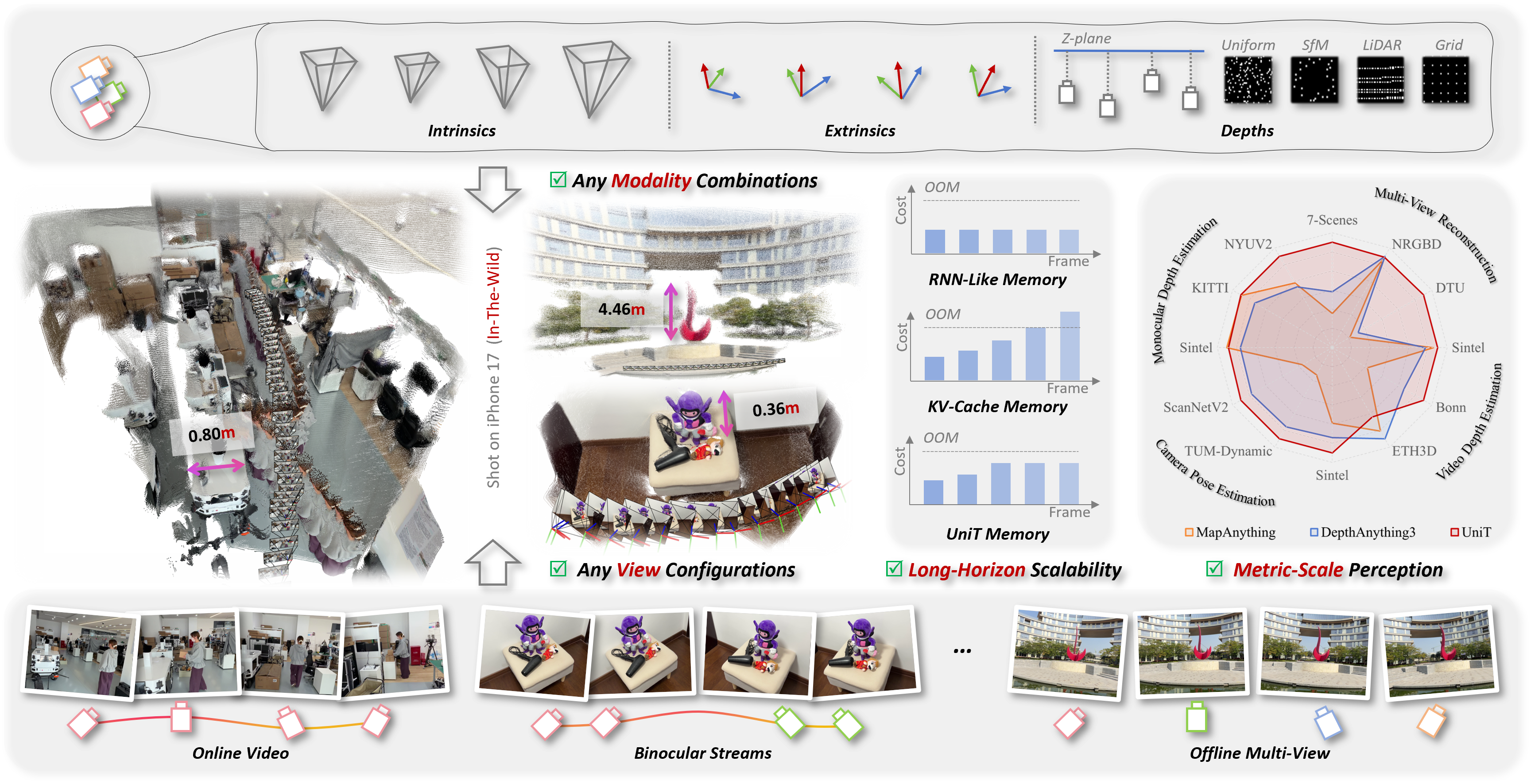}\caption{
    We present \textsc{UniT}, a feed-forward model for unified geometry learning. It supports a broad range of essential geometry perception capabilities across diverse \textit{view configurations}, \textit{modality combinations}, \textit{metric-scale perception}, and \textit{long-horizon scalability}. Specifically, it enables both online and offline inference over an arbitrary number of views, flexibly incorporates auxiliary modalities such as camera parameters and depth maps, recovers geometry at metric scale in physical units, and maintains bounded memory over long horizons in in-the-wild environments.
}

}

\begin{document}

\title{\textsc{UniT}: Unified Geometry Learning with Group Causal Transformer}

\author{Haotian~Wang\textsuperscript{1,2} \and
Yusong~Huang\textsuperscript{1} \and
Zhaonian~Kuang\textsuperscript{2,1} \and
Hongliang~Lu\textsuperscript{1} \and
Xinhu~Zheng\textsuperscript{1} \and
Meng~Yang\textsuperscript{2} \and
Gang~Hua\textsuperscript{3}}
\institute{%
\ijcvcorrespondence{Xinhu Zheng}{xinhuzheng@hkust-gz.edu.cn}
\ijcvinstitutegap{5pt}
\ijcvcorrespondence{Meng Yang}{mengyang@mail.xjtu.edu.cn}
\ijcvinstitutegap{10pt}
\ijcvaffiliation{1}{Intelligent Transportation Thrust of the Systems Hub, The Hong Kong University of Science and Technology (GZ), Guangzhou, P.R.China.}
\ijcvinstitutegap{5pt}
\ijcvaffiliation{2}{National Key Laboratory of Human-Machine Hybrid Augmented Intelligence, Institute of Artificial Intelligence and Robotics, Xi'an Jiaotong University, Xi'an, P.R.China.}
\ijcvinstitutegap{5pt}
\ijcvaffiliation{3}{Applied Science, Amazon.com, Inc., USA.}}

\abstract{
\rev{Recent feed-forward geometry perception models have made substantial progress in inferring dense 3D structure from sensor observations.
However, this progress has largely been driven by task-specific designs, leaving essential capabilities such as online perception, offline reconstruction, long-horizon scalability, multi-modal integration, and metric-scale estimation fragmented across multiple incompatible paradigms.
We present \textsc{UniT}, a novel \textit{Group Causal Transformer} that unifies these seemingly disparate capabilities within a single formulation to enable unified geometry learning.
Our first insight is to formulate geometry learning as \textit{``next-group-prediction"}, allowing the same model to operate online through successive steps over single-frame groups or offline by processing a multi-frame group in a single step.
To make this formulation scalable over long horizons and effective for metric-scale generalization, we further develop \textit{queue-style KV caching} and a \textit{scale-adaptive geometry loss}, respectively.
The former leverages anchor-free relational modeling to reduce long-range dependencies on early frames, allowing outdated cached states to be safely discarded for bounded memory usage.
The latter factorizes the metric-scale assumption into scale-invariant constraints and a partial metric-scale term, implicitly encouraging a progressive transition from learning easier scale-invariant geometry to recovering more challenging metric scale.
Together with a plug-in \textit{modal attention} module for auxiliary modalities, \textsc{UniT} achieves state-of-the-art performance in unified geometry perception, as validated on eleven benchmarks spanning seven representative tasks against nine strong baselines.}

}
\keywords{
3D, Point map, Multi-modal, Metric-scale, Streaming, Unified model.
}

\projectpage{https://omega-hkustgz.github.io/projects/unit}
\maketitle
\begin{figure*}[t]
\centering
\insertfig
\end{figure*}

\section{Introduction}
Geometry perception, the task of inferring dense 3D structure from sensor observations, plays a substantial role in a wide range of applications, including robotics\cite{kim2024openvla}, augmented reality\cite{arena2022overview}, and autonomous systems\cite{chen2024end}. Driven by their remarkable robustness and efficiency, recent advances have shifted the field from optimization-based pipelines such as Structure-from-Motion (SfM) \cite{schoenberger2016sfm} and Simultaneous Localization and Mapping (SLAM) \cite{campos2021orb} toward feed-forward models built upon the point map representation \cite{wang2024dust3r}.

\begin{figure}[!t]
    \centering
    \includegraphics[width=\linewidth]{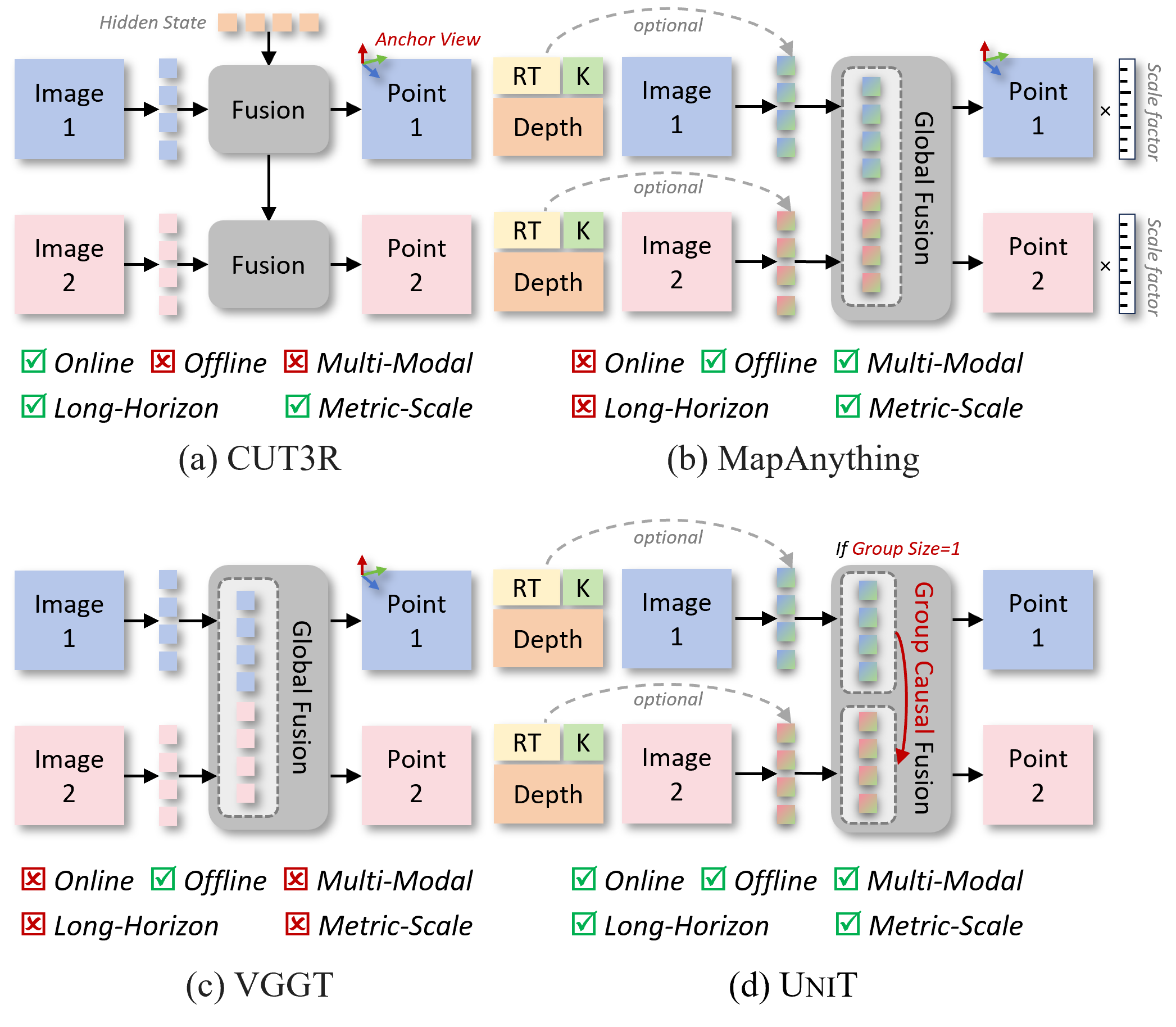}
    \caption{Four representative paradigms for geometry perception: (a) CUT3R \cite{wang2025continuous}, (b) MapAnything \cite{keetha2025mapanything}, (c) VGGT \cite{wang2025vggt}, and (d) our \textsc{UniT}.}
    \label{fig:paradigm}
\end{figure}

\rev{Despite rapid progress, these feed-forward models have largely
been driven by task-specific designs. As shown in \cref{fig:paradigm}, five essential capabilities of geometry perception are distributed across largely incompatible paradigms: (a) \textit{online} sequential inference for continuous perception \cite{zhuo2025streaming}, (b) \textit{offline} joint reconstruction from accumulated observations \cite{wang2025pi}, (c) \textit{multi-modal} fusion for flexible sensor integration \cite{jang2025pow3r}, (d) \textit{long-horizon} scalability for extended spatiotemporal reasoning \cite{chen2025ttt3r}, and (e) \textit{metric-scale} estimation for physically grounded geometry \cite{lin2025depthany3}. This fragmentation limits their applicability across diverse downstream platforms.}

\rev{The reason lies in fundamentally different formulations across geometry learning paradigms. For example, CUT3R \cite{wang2025continuous} formulates streaming perception as frame-by-frame prediction, decoding one point map per step over long horizons, as illustrated in \cref{fig:paradigm}~(a). In contrast, VGGT \cite{wang2025vggt} formulates offline reconstruction as joint inference over all frames, decoding all point maps within a single forward pass, as shown in \cref{fig:paradigm}~(c). MapAnything \cite{keetha2025mapanything} further extends offline paradigm by introducing separate components for multi-modal integration and metric-scale estimation, as shown in \cref{fig:paradigm}~(b). These specialized formulations entangle model capabilities with target applications, making feed-forward models to evolve along fragmented paths.}

\rev{In this paper, we aim to show that these seemingly disparate challenges can be addressed within a unified formulation, the proposed \textbf{Group Causal Transformer}. Along the pathway toward this unified geometry learning, we identify and address three key challenges.}

\subsection{Incompatible assumptions on view configurations}

Online methods incrementally update geometry over time, while offline methods reconstruct the entire scene jointly within a single step. This fundamental discrepancy renders online methods inefficient for multi-step aggregation in offline scenarios \cite{wang2025pi}, while offline methods incur redundant recomputation whenever new frames arrive in streaming settings \cite{zhuo2025streaming}.

\rev{In this work, we reveal that these seemingly heterogeneous view configurations can be formulated as \textit{``next-group-prediction"}. By varying the group size to control the number of frames jointly processed in each forward pass, the model can seamlessly transition between different inference behaviors within the same group causal process.}

As illustrated in \cref{fig:paradigm}~(d), our group causal process employs bidirectional attention \cite{devlin2019bert} within each group and causal attention \cite{achiam2023gpt} across groups. When the group size is set to one, this formulation naturally reduces to an online pipeline with sequential processing over time. At the other extreme, when the group size spans the full sequence, it degenerates into an offline architecture without temporal causality.

Beyond the standard online and offline modes, this formulation additionally accommodates multi-camera array streams, which are commonly employed in robotics and autonomous driving \cite{li2024bevformer}. In such scenarios, group sizes typically range from four to eight, enabling joint reasoning over multiple synchronized views.

\subsection{Unbounded growth of streaming memory}

In streaming architectures, historical information is stored as KV-cache entries accumulated from the first frame up to the current step \cite{zhuo2025streaming}. Consequently, both memory and computational costs grow with sequence length, making long-horizon inference inefficient and limiting its scalability \cite{yuan2026infinitevggt}. 

\rev{In this work, we show that reducing long-range dependencies on early frames enables bounded memory inference over extended horizons, without relying on complicated memory compression techniques  \cite{yuan2026infinitevggt,wang2025flashvggt,shen2025fastvggt}. By modeling only relative relations among available viewpoints, our approach allows a \textit{queue-style KV caching} mechanism to safely discard outdated cached states.}

\rev{More specifically, such long-range dependencies largely arise from the fixed first-frame reference adopted by most existing methods \cite{zhuo2025streaming,wu2025point3r}. This design expresses every frame relative to the initial anchor, thus encouraging persistent reliance on distant past frames.}

\rev{Motivated by this, we integrate the anchor-free techniques \cite{wang2025pi} into our group causal modeling, allowing KV-cache to be managed as a fixed-capacity queue of size $Q$. As a result, the per-step computational complexity is bounded by $O(Q)$, instead of scaling linearly with sequence length.}

\subsection{Limited generalization in metric-scale learning} 

Due to inherent scale ambiguity \cite{ranftl2020midas}, recovering metric scale is substantially more challenging than learning relative geometry. The former involves a large dynamic range and exhibits weaker cross-scene generalization \cite{wang2025moge2}, making it a long-standing challenge in geometry perception \cite{piccinelli2025unidepthv2}.

\rev{In this work, we show that factorizing the metric-scale assumption into scale-invariant constraints \cite{wang2024dust3r} and a partial metric-scale term can effectively improve cross-scene generalization. The resulting \textit{scale-adaptive geometry loss} implicitly encourages a progressive transition from learning easier relative structure to recovering more challenging absolute scale, thereby improving training stability.}

\rev{Unlike approaches that estimate an explicit scale factor separately \cite{wang2025moge2,keetha2025mapanything}, our method incorporates metric-scale supervision directly into the geometry objective. This formulation admits a closed-form solution for metric scale, avoiding an extra scale-estimation module while preserving end-to-end geometry learning.}

\rev{Together with a plug-in \textit{modal attention} module for flexible integration of auxiliary modalities, we finally arrive at our group causal transformer, namely \textbf{\textsc{UniT}}, that effectively unifies the five essential capabilities within a single formulation.}

Extensive experiments on eleven benchmark datasets validate the effectiveness of \textsc{UniT} across a wide range of view configurations, modality combinations, scale conditions, and sequence lengths. The validation covers seven representative tasks including multi-view reconstruction, camera pose estimation, video depth estimation, monocular depth estimation, long-horizon perception, multi-modal reconstruction, and depth completion. The results show that \textsc{UniT} achieves state-of-the-art performance in unified geometry perception.

\rev{In summary, our contributions are threefold:}
\begin{enumerate}
    \item \rev{We introduce \textsc{UniT}, a novel \textit{Group Causal Transformer}, to enable unified geometry learning. It supports arbitrary view configurations and modality combinations while allowing long-horizon scalability and metric-scale perception within a single formulation.}
    
    \item \rev{We explore three key challenges in unified geometry learning: \textit{``next-group-prediction''} provides a unified formulation for diverse view configurations; \textit{queue-style KV caching} maintains bounded memory over long horizons; and \textit{scale-adaptive geometry loss} improves metric-scale generalization across scenes.}
    
    \item \rev{Extensive experiments on eleven benchmarks spanning seven representative tasks demonstrate \textsc{UniT}'s state-of-the-art performance in unified geometry perception.}
\end{enumerate}

\section{Related Work}
\subsection{Offline Geometry Perception} 
Following the success of DUSt3R\cite{wang2024dust3r}, a series of feed-forward methods of geometry perception have emerged based on the point map representation, supporting a range of tasks such as multi-view reconstruction\cite{schoenberger2016sfm}, camera pose estimation\cite{nister2004visual}, and video \cite{hu2025depthcrafter} and monocular depth estimation\cite{eigen2014depth}. This representation unifies 2D-to-3D correspondence learning and 3D-to-3D geometric reasoning within a single representation, enabling effective end-to-end reconstruction from unconstrained image pairs. However, DUSt3R was limited to processing only two images per forward pass, which led to iterative computational overhead and expensive global alignment procedures when extended to longer image sequences. To alleviate this limitation, the MASt3R line of works \cite{leroy2024grounding,murai2025mast3r,duisterhof2025mast3r} revisited key principles from classical multi-view geometry, such as correspondence matching and graph-based view relationships, to better leverage optimization-inspired advantages in multi-view settings. 

More broadly, recent methods such as Fast3R\cite{yang2025fast3r} and VGGT\cite{wang2025vggt} introduced transformer-based parallel processing modules that enabled multiple viewpoints to be processed within a single forward pass, substantially reducing computational complexity while improving performance in multi-view scenarios. These advances have strongly motivated the community of geometry perception, leading to the emergence of more 3D foundation models, such as $\pi^3$\cite{wang2025pi} and DepthAnything3\cite{lin2025depthany3}. In particular, $\pi^3$ highlighted the limitation of the fixed reference view and proposes an anchor-free camera loss to alleviate it. Despite their strong performance in offline settings, these methods assume fully observed inputs and lack support for incremental or long-horizon inference.

\subsection{Online Geometry Perception}
To support real-time applications with streaming observations, such as robotics and autonomous driving, recent studies have investigated incremental reasoning strategies for online 3D scene perception. In contrast to pair-based methods \cite{wang2024dust3r,fang2026dens3r} and offline methods\cite{wang2025vggt,gao2026more}, Spann3R\cite{wang20253d} and CUT3R\cite{wang2025continuous} employed recurrent-style frameworks that maintain a constant-sized hidden state as spatial memory. At each time step, the model sequentially incorporated a new image observation, updated the spatial memory, and predicted the corresponding point map. These strategies achieve high computational efficiency over time, facilitating real-time deployment and long-horizon perception.

To further alleviate forgetting in long sequences, Point3R\cite{wu2025point3r} adopted an explicit memory design that stores historical image tokens to anchor the global coordinate system robustly. Compared to the constant-sized memory of CUT3R, Point3R expanded its memory capacity over time, resulting in increased computational overhead. In a complementary direction, TTT3R\cite{chen2025ttt3r} further extended CUT3R with a test-time learning paradigm, dynamically updating hidden states via a confidence-guided integration of historical memory and new observations. StreamVGGT\cite{zhuo2025streaming} represented another research direction, introducing KV-cache-based memory following the causal formulation. However, StreamVGGT relied on all historical KV entries, thereby limiting scalability in the long-horizon setting \cite{yuan2026infinitevggt}.

\subsection{Geometry Perception Extensions} Beyond the view configurations considered by offline and online methods, extensive efforts have been devoted to broader capabilities, including multi-modal integration, metric-scale estimation, and long-horizon perception.

In the multi-modal setting, an early exploration is Pow3R\cite{jang2025pow3r}. It extended DUSt3R by incorporating auxiliary modalities, such as camera intrinsics, extrinsics, and depth maps, as optional conditions embedded into image tokens. Inspired by this design, many offline\cite{liu2025worldmirror,peng2025omnivggt,keetha2025mapanything,lin2025depthany3} and online\cite{khafizov2025g} approaches have adopted plugin-based architectures to flexibly integrate additional geometric cues. Among them, MapAnything\cite{keetha2025mapanything} stands out as a representative framework that unifies multi-modal inputs and metric-scale estimation within a single model through a factored representation. DepthAnything3\cite{lin2025depthany3} also supported metric-scale prediction and incorporates camera parameters in a nested manner.

For long-horizon perception, VGGT-Long\cite{deng2025vggt} decomposed the extended trajectories into multiple overlapping short sequences and subsequently realigned them to enable kilometer-scale reconstruction, albeit at the cost of substantial redundant computation. In parallel, several studies have investigated memory compression strategies, such as token merging strategies\cite{shen2025fastvggt}, compact spatial descriptors\cite{wang2025flashvggt}, and token updating strategies\cite{yuan2026infinitevggt}. While these methods considerably broaden the applicability of feed-forward models, they primarily focus on memory compression. In contrast, \textsc{UniT} reduces long-range dependencies on early frames through a simple queue-style KV caching mechanism, making it orthogonal to existing methods and readily compatible with them.

\section{Method}

\begin{figure*}[!t]
    \centering
    \includegraphics[width=0.98\linewidth]{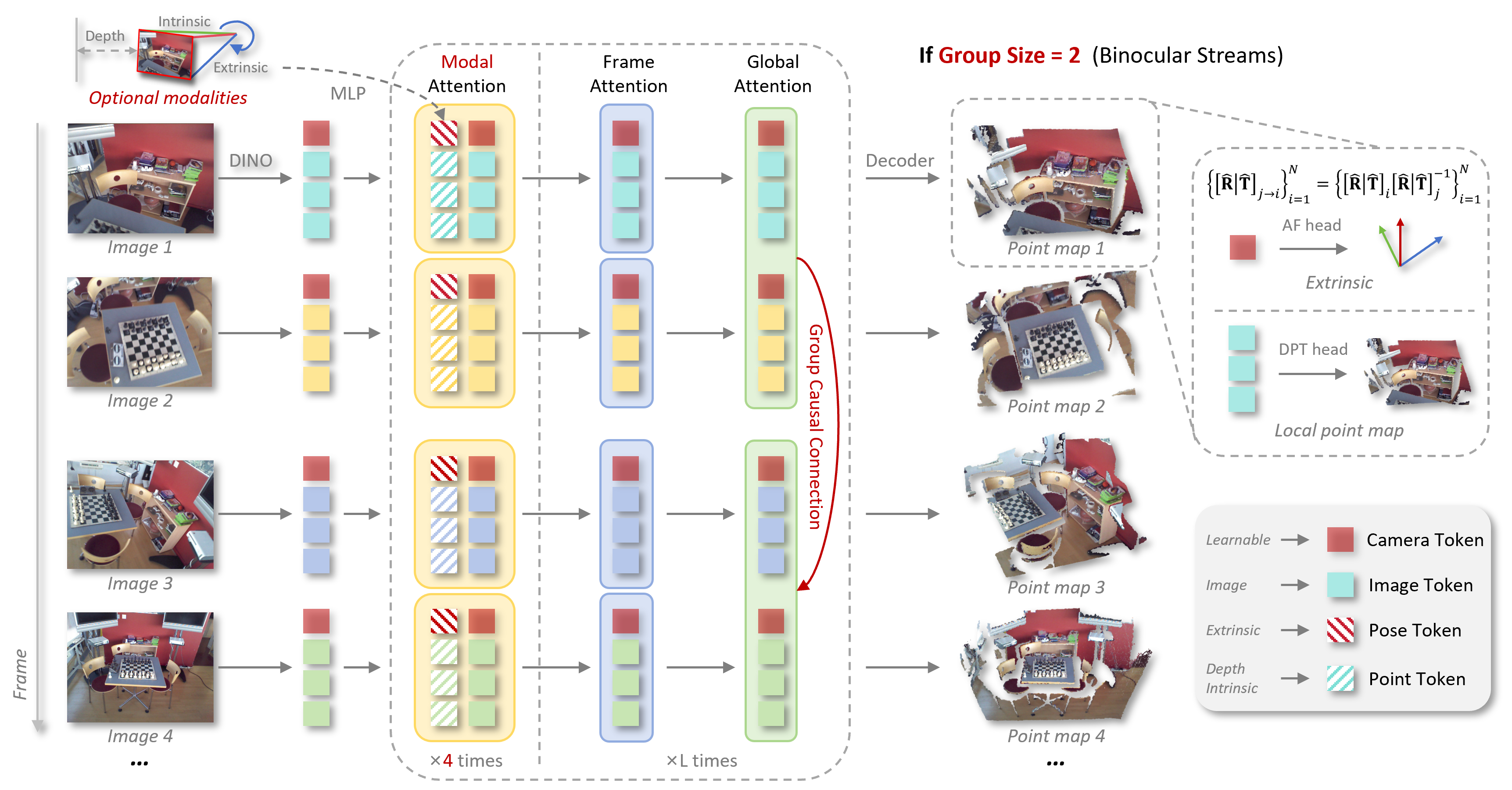}
    \caption{Architecture Overview. Image groups are first patchified into tokens using DINO \cite{oquab2023dinov2}. These tokens are then fused with tokens encoded from optional modalities through a modal attention layer, followed by frame attention and global attention layers. In global attention, bidirectional attention operates within each group, while causal attention is applied across groups. The fused tokens are finally decoded into global point maps, represented by a local point map using DPT head \cite{ranftl2021vision} and camera extrinsics with our anchor-free (AF) camera head. To control model complexity, modal attention is applied four times.}
    \label{fig:framework}
\end{figure*}

\subsection{Group Causal Formulation}
\label{sec:formulation}
The goal of geometry perception is to predict a sequence of target point maps $\{\mathbf{X}_t\}_{t=1}^{N}$ from image observations $\{\mathbf{I}_t\}_{t=1}^{N}$ with sequence length $N$. Beyond RGB images, we aim to flexibly support multi-modal inputs that may be available in real-world scenarios, including depth maps $\{\mathbf{D}_t\}_{t=1}^{N}$, camera intrinsics $\{\mathbf{K}_t\}_{t=1}^{N}$, and camera extrinsics $\{[\mathbf{R}|\mathbf{T}]_t\}_{t=1}^{N}$, where $\mathbf{R}\!\in\!\mathrm{SO}(3)$ and $\mathbf{T} \!\in \!\mathbb{R}^3$ denote the rotation matrix and translation vector, respectively. Formally, geometry perception is modeled as a conditional distribution:
\begin{equation}
p\left( \{\mathbf{X}_t\}_{t=1}^{N} \,\middle|\, 
\{\mathbf{I}_t\}_{t=1}^{N},
\{\mathbf{\mathcal{O}}_t\}_{t=1}^{N}\right),
\label{eq:basic}
\end{equation}
where $\mathbf{\mathcal{O}}_t \subseteq \{\mathbf{D}_t, \mathbf{K}_t, [\mathbf{R}|\mathbf{T}]_t\}$ denotes an optional subset of multi-modal signals at time $t$. 

\noindent \textbf{Causal Modeling.}
The joint conditional distribution naturally admits a causal factorization:
\begin{equation}
p\left( \{\mathbf{X}_t\}_{t=1}^{N} \,\middle|\, 
\{\mathbf{I}_t\}_{t=1}^{N}, 
\{\mathbf{\mathcal{O}}_t\}_{t=1}^{N}\right)
=\prod_{t=1}^{N} p\!\left( \mathbf{X}_t \,\middle|\, \mathbf{I}_{\le t}, \mathbf{\mathcal{O}}_{\le t}\right)
\label{eq:autoreg}
\end{equation}
where $\mathbf{I}_{\le t}$ denotes $\{\mathbf{I}_{\tau}\}_{\tau=1}^{t}$, which represents the past and current image observations up to time $t$ for predicting $\mathbf{X}_t$. Based on this causal formulation, the target point maps $\{\mathbf{X}_t\}_{t=1}^{N}$ are estimated by maximizing the conditional likelihood with model $\Theta$ in an sequential manner:
\begin{equation}
\{\mathbf{X}_t\}_{t=1}^{N}
\leftarrow \arg\max_{\Theta}
\prod_{t=1}^{N}
p\left(\mathbf{X}_t\,\middle| \mathbf{I}_{\le t}, \mathbf{\mathcal{O}}_{\le t}\right),
\label{eq:autoregressive}
\end{equation}

This causal formulation describes an online inference process driven by ``next-frame-prediction" \cite{zhuo2025streaming}, where the point map $\mathbf{X}_t$ is predicted sequentially at each time step $t$. The accumulated predictions thus result in the full target sequence $\{\mathbf{X}_t\}_{t=1}^{N}$. 


\noindent \textbf{Group Causal Modeling.}
\rev{To unify different view configurations within a single formulation, we generalize the causal process in \cref{eq:autoregressive} to a ``next-group-prediction" formulation.} Specifically, at each time step $t$, a group of point maps $\mathbf{X}_t^{1:G}$ is treated as a single prediction unit, where $G$ denotes the number of viewpoints jointly observed at that time step. The resulting group causal modeling is formally defined as
\begin{equation}
\{\mathbf{X}_t^{1:G}\}_{t=1}^{N/G}\
\leftarrow \arg\max_{\Theta}
\prod_{t=1}^{N/G}
p\left(\mathbf{X}_t^{1:G} \middle| \mathbf{I}_{\le t}^{1:G}, \mathbf{\mathcal{O}}_{\le t}^{1:G}\right),
\label{eq:group_autoregressive}
\end{equation}
When $G\!=\!1$, the formulation reduces to standard \textit{online} inference with a sequential process, as shown in \cref{eq:autoregressive}. When $G\!=\!N$, it reduces to a single-step inference process, recovering the \textit{offline} parallel setting without temporal dependency. As $G$ varies from 1 to $N$, the formulation naturally unifies diverse view configurations, ranging from monocular video to multi-view reconstruction. An example of binocular streaming with $G\!=\!2$ is illustrated in \cref{fig:framework}.

\subsection{Group Causal Transformer}
\label{sec:transformer}
\rev{To instantiate the proposed group causal formulation in \cref{eq:group_autoregressive}}, we build our group causal transformer upon the Visual Geometry Grounding Transformer (VGGT) \cite{wang2025vggt}. The overall architecture is illustrated in \cref{fig:framework}.

\begin{figure}[!t]
    \centering
    \includegraphics[width=\linewidth]{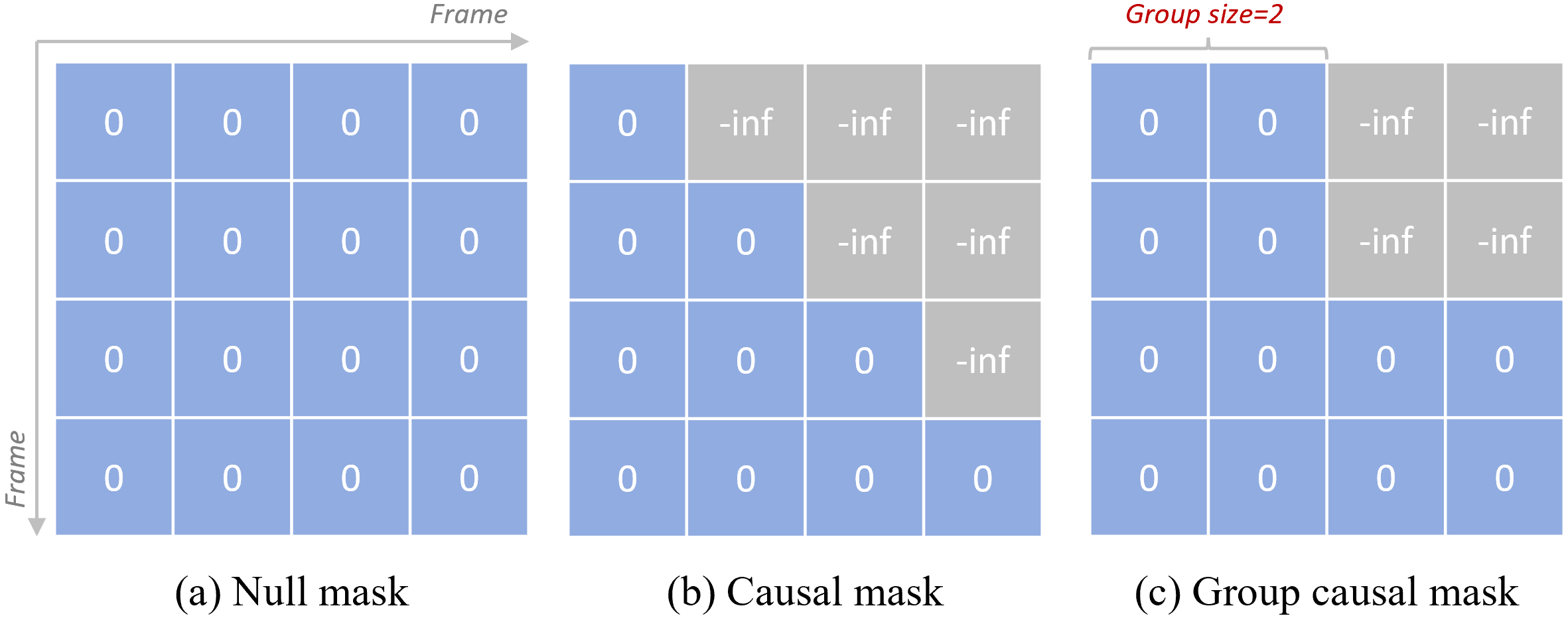}
    \caption{Illustration of three types of attention mask, including (a) null mask, (b) standard causal mask, and (c) our group causal mask with a group size of 2. Entries marked as ``-inf" indicate masked tokens. For simplicity, each token denotes one frame in this illustration.}
    \label{fig:mask}
\end{figure}

\noindent \textbf{Visual Geometry Grounding Transformer.} VGGT presents a concise architecture for geometry perception in image-only, offline settings. It first extracts image tokens from visual observations $\{\mathbf{I}_t\}_{t=1}^{N}$ by DINO\cite{oquab2023dinov2}, and then processes them through $L$ layers of alternating attention. Specifically, each alternating attention layer consists of a frame attention that independently models intra-frame relationships, followed by a global attention that captures interactions across all frames. This process can be written as
\begin{equation}
\mathbf{H}_t = \text{Attn}\!\left(\{\mathbf{F}_t\}_{t=1}^{N}\right), \mathbf{F}_t=\text{Attn}\big(\text{DINO}(\mathbf{I}_t)\big),
\label{eq:aaattention}
\end{equation}
where $\mathbf{H}_t$, $\mathbf{F}_t$ denote the resulting feature tokens from global and frame attentions for frame $t$. Finally, multiple redundant predictions, such as point maps, depth maps, camera parameters, and keypoint tracking, are decoded from these feature tokens using different heads.

\noindent \textbf{Group Causal Transformer.} Based on the group causal formulation in \cref{eq:group_autoregressive}, the original attention block in \cref{eq:aaattention} is modified in three aspects:
\begin{enumerate}
    \item \textit{Temporal Causality}: Temporal causality should be introduced into the global attention, where the model only attends to observations $\mathbf{I}_{\le t}$ up to time step $t$;
    \item \textit{Group-wise Unit}: The prediction unit is defined as a group of observations $\mathbf{I}_t^{1:G}$, which are processed with bidirectional attention at time step $t$;
    \item \textit{Auxiliary Modalities}: Auxiliary signals $\mathbf{\mathcal{O}}_{t}$ at time step $t$ are incorporated as flexible multi-modal conditions.
\end{enumerate}

Accordingly, the proposed group causal transformer reformulates the alternating attention layers as
\begin{equation}
\ddot{\mathbf{H}}_t = \text{Attn}\!\left({\ddot{\mathbf{F}}}_{\le t}^{1:G}\right), \qquad
\ddot{\mathbf{F}}_t^{g} = \text{Attn}\!\left(\mathbf{M}_t^{g}\right),
\label{eq:updated_global}
\end{equation}
where $\ddot{\mathbf{H}}_t$, $\ddot{\mathbf{F}}_t$ denote feature tokens of the updated global and frame attentions, respectively. $\mathbf{M}_t^{g}$ denotes the fused tokens from the image $\mathbf{I}_t^{g}$ and multi-modal signals $\mathbf{\mathcal{O}}_{t}^g$ at time $t$ with group index $g\! \in \!\{1,\ldots,G\}$, obtained via the proposed \textit{Modal Attention} layer,
\begin{equation}
\mathbf{M}_t^{g} = \text{ModalAttn}\big(\text{DINO}(\mathbf{I}_t^{g}),\, \text{MLP}(\mathbf{\mathcal{O}}_{t}^{g})\big).
\label{eq:crossattn}
\end{equation}
In the following, we introduce the group causal connection $\ddot{\mathbf{F}}_{\le t}^{1:G}$ in the global attention layer, as well as the architecture of the modal attention $\text{ModalAttn}$.

\noindent \textbf{Group Causal Connection.} In modern streaming transformers \cite{achiam2023gpt,touvron2023llama}, causal dependencies are typically implemented by applying causal masks within attention layers, which prevent future observations from influencing the current prediction. As shown in \cref{fig:mask}~(b), the standard causal mask assigns negative infinity to future positions, thereby disabling attention to these tokens \cite{vaswani2017attention}.

To implement the group causal connection defined in \cref{eq:updated_global}, we associate each time step with a group of observations and enforce causality at the group level. Specifically, bidirectional attention is performed within each group, while causal attention is applied across groups. 

\rev{This design can be implemented by simply modifying the standard causal attention mask.} An example of attention mask with $G\!=\!2$ is illustrated in \cref{fig:mask}~(c), where tokens from future groups are masked out. When $G$ varies from $1$ to $N$, this group causal mask allows the model to handle arbitrary view configurations with multiple synchronized cameras.

\noindent \textbf{Modal Attention.} In our framework, the optional multi-modal inputs $\mathbf{\mathcal{O}}_{t}^{g}$ are first encoded by a two-layer MLP with SP-Normalization \cite{wang2024scale}, where absent modalities are represented as $\mathbf{0}$ matrices. This yields two complementary types of modal tokens. The first type, point tokens, provides a dense representation by encoding depth maps together with local ray maps derived from camera intrinsics. Compared with compact intrinsic parameters, local ray maps retain pixel-wise coordinates and therefore capture richer spatial cues. The second type, pose tokens, offers a compact parametric representation by encoding the 12D camera extrinsics.

\begin{figure}[!t]
    \centering
    \includegraphics[width=\linewidth]{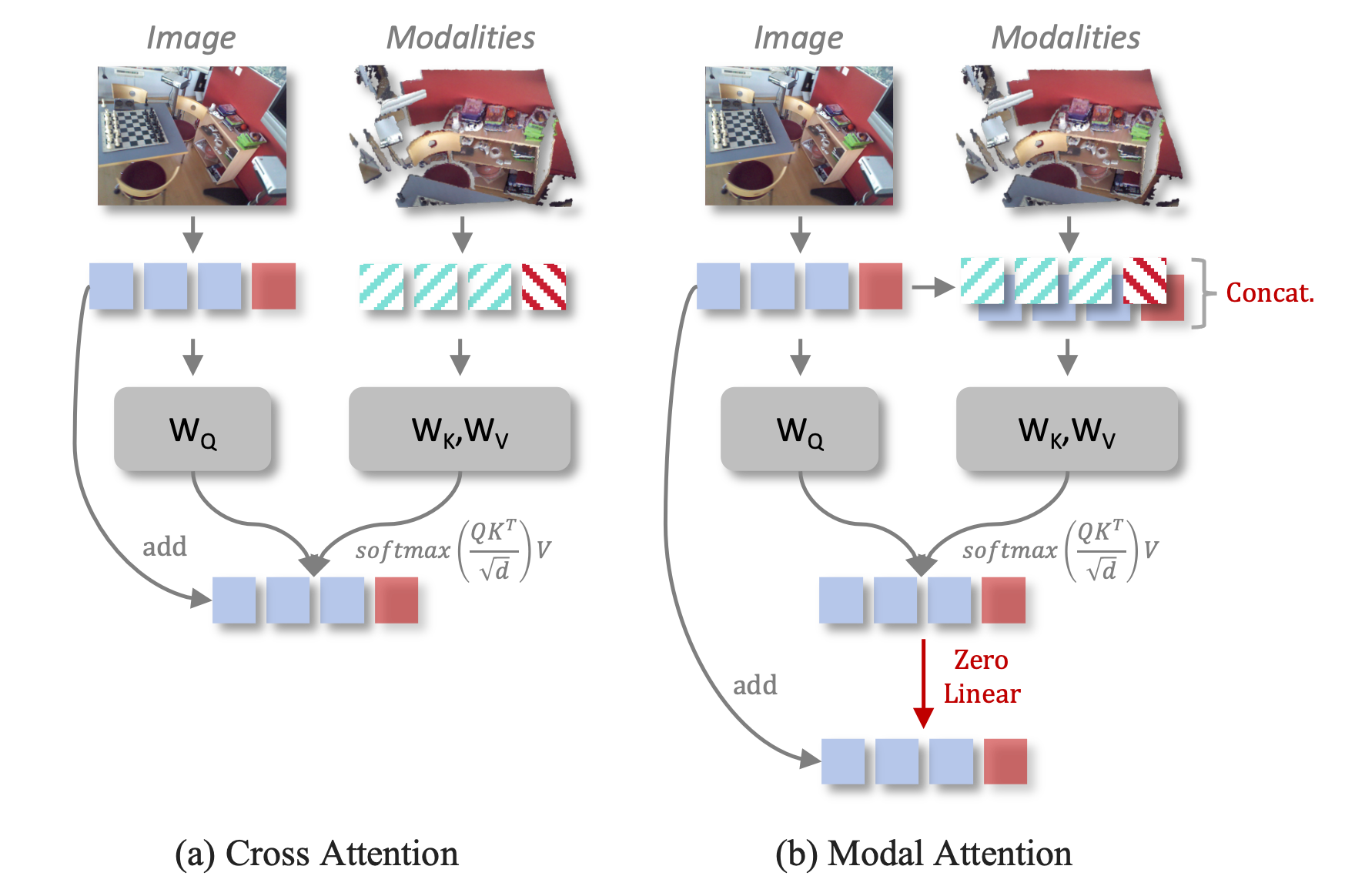}
    \caption{Illustration of two types of attention layers, including (a) standard cross attention and (b) the proposed modal attention.}
    \label{fig:crossattn}
\end{figure}

Then, the modal tokens are fused with image tokens through modal attention layers, as illustrated in \cref{fig:crossattn}~(b). \rev{The layer is adapted from the standard cross-attention shown in \cref{fig:crossattn}~(a), but employing two modifications.} First, we concatenate image and modal tokens at spatially aligned positions before attention. This design explicitly encodes pixel-wise correspondences between modalities, enabling more spatially aware multi-modal interactions. Second, we add a zero-initialized linear projection layer \cite{lin2025prompting}, allowing the model to effectively inherit the pretrained knowledge from VGGT.

To control model complexity, modal attention is inserted at four stages following the stage partition of the DPT head \cite{ranftl2021vision} used in VGGT, specifically at layers [0, 5, 12, 18] when $L\!=\!24$. These modules account for only about 3\% of the total parameters. The effectiveness of these design choices is validated through ablation studies in \cref{sec:ablation}.

We next describe two key components of the group causal transformer, which are introduced for efficient long-horizon scalability in \cref{sec:queuekv} and metric-scale generalization in \cref{sec:metricscale}, respectively.

\subsection{Queue-Style KV Caching}
\label{sec:queuekv}
In streaming transformers, the KV-cache serves as a memory context that enables efficient inference for streaming inputs such as $G\!<\!N$. However, its memory footprint grows unboundedly with sequence length \cite{zhuo2025streaming}. Existing methods primarily address this issue through cache compression \cite{yuan2026infinitevggt,wang2025flashvggt,shen2025fastvggt}, aiming to reduce the number of cached tokens. \rev{We instead explore an orthogonal direction that reduces the reliance on early observations, making outdated KV-cache entries can be safely discarded once a predefined cache capacity is reached.}

\begin{figure}[!t]
    \centering
    \includegraphics[width=\linewidth]{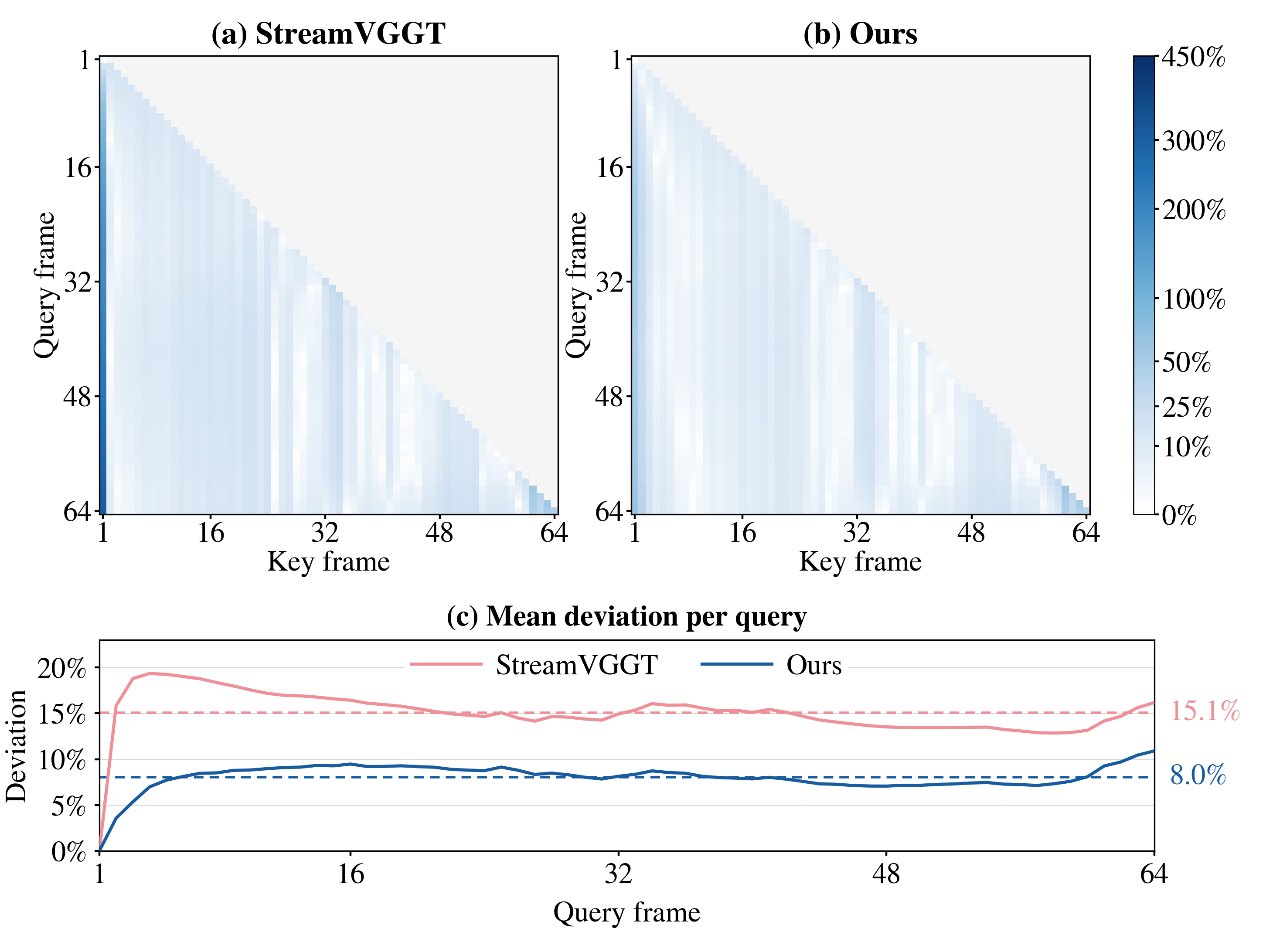}
    \caption{\rev{Global attention distribution in the streaming mode. The anchor-based method (a) StreamVGGT concentrates more attention on early frames than (b) our anchor-free method. The curves in (c) show that our method distributes attention more uniformly across frames, with lower attention deviation.}}
    \label{fig:attn_dis}
\end{figure}

\rev{As shown in \cref{fig:attn_dis}, we attribute this to the fixed first-frame reference commonly adopted by feed-forward models. Since each frame is expressed relative to the initial anchor, the model is encouraged to maintain persistent dependence on distant past frames. Motivated by this, we integrate the anchor-free techniques \cite{wang2025pi} into our group causal modeling to remove this dependency, which enables our KV-cache to be maintained as a fixed-capacity queue with bounded memory.}

\noindent \textbf{Anchor-Free Extrinsic Loss.} \rev{To remove the long-range dependencies on early frames, we adopt an anchor-free extrinsic loss introduced in $\pi^3$ \cite{wang2025pi} that enforces pairwise consistency among camera extrinsics with relative modeling.} Specifically, given the predicted extrinsic $[\mathbf{\hat{R}}|\mathbf{\hat{T}}]_i$ and its ground-truth counterpart $[\mathbf{R}|\mathbf{T}]_i$, the loss is defined as follows:
\begin{equation}
\mathcal{L}_{rel}^{cam}=\frac{1}{N(N-1)}\sum_{i \neq j}\left(\nabla_{rot}(i,j)+\lambda \nabla_{trans}(i,j)\right).
\label{eq:loss_rel_cam}
\end{equation}
Here, 
$\nabla_{rot}(i,j)
\!=\! \arccos\left( 
\left(\text{Tr}\left( \mathbf{\hat{R}}_{j \rightarrow i}^{-1} \mathbf{R}_{j \rightarrow i}\right) \!-\!1 \right)/2
\right)$ and $\nabla_{trans}(i,j)\!=\!
\left\| \mathbf{\hat{T}}_{j \rightarrow i}/\hat{s} \!-\! \mathbf{T}_{j \rightarrow i}/s \right\|_1$. $\mathbf{R}_{j \rightarrow i}$ and $\mathbf{T}_{j \rightarrow i}$ denote the relative rotation and translation from view $j$ to view $i$. $\hat{s}$ and $s$ denote the global scale factors. In our implementation, they are computed using the $\ell_2$ norm on predicted depth maps $\{\mathbf{\hat{D}}_i\}_{i=1}^{N}$ and ground-truth ones $\{\mathbf{D}_i\}_{i=1}^{N}$ over the entire sequence, following prior work \cite{wang2024dust3r,deng2025vggt,wang2025continuous}. The scalar $\lambda$ is a weighting hyperparameter, set to $\lambda = 10$.

\begin{figure}[!t]
    \centering
    \includegraphics[width=\linewidth]{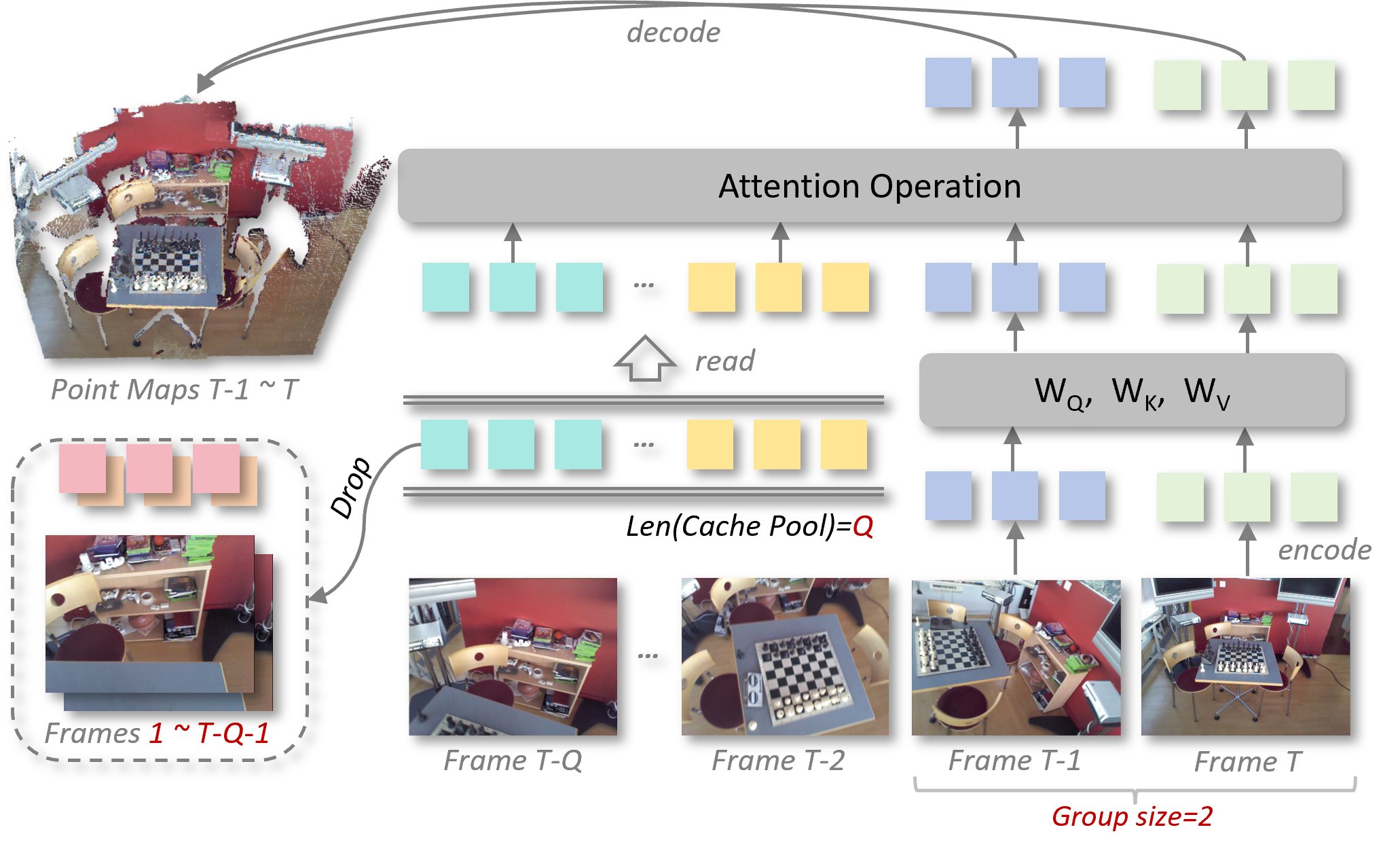}
    \caption{Illustration of queue-style KV caching. The KV-cache is organized as a fixed-length queue, in which outdated tokens are dropped once the predefined capacity $Q$ is exceeded.}
    \label{fig:cache}
\end{figure}

\noindent \textbf{Anchor-Free Camera Head.} To further extend the anchor-free design to point map constraints, we redesign the camera head to re-parameterize poses via relative transformations:
\begin{equation}
\{[\mathbf{\hat{R}}|\mathbf{\hat{T}}]_{j\rightarrow i}\}_{i=1}^{N}=
\{[\mathbf{\hat{R}}|\mathbf{\hat{T}}]_i{[\mathbf{\hat{R}}|\mathbf{\hat{T}}]}_j^{-1}\}_{i=1}^{N}. 
\label{eq:cam_head}
\end{equation}
This design makes the pose representation invariant to any shared global transformation, so that it encodes only the relative relationships among views. Accordingly, the point map prediction $\mathbf{\hat{X}}_{j\rightarrow i}$, parameterized by the relative pose $[\mathbf{\hat{R}}|\mathbf{\hat{T}}]_{j\rightarrow i}$, is defined in the same relative coordinate system.

When the same re-parameterization is applied to the ground-truth poses, regular point map losses can likewise be defined in this anchor-free system. For notational simplicity, we hereafter omit the explicit re-parameterization notation $j\!\rightarrow\!i$ and use the subscript $i$ instead.

\noindent \textbf{Queue-Style KV Caching.} Benefiting from anchor-free causal modeling, geometric relationships are represented through relative transformations, allowing each KV cache entry to be stored independently without relying on early frames. This property enables the KV-cache to be maintained as a fixed length queue, where outdated caches are discarded once a predefined queue length $Q$ is exceeded. \cref{fig:cache} illustrates our queue-style KV caching mechanism.

The queue-style KV caching ensures constant computational complexity per step, independent of the sequence length $N$. Specifically, naive global attention incurs quadratic complexity $\mathcal{O}(N\!\times\!N)$. By contrast, KV caching avoids recomputing keys and values of attention for previous time steps, resulting in linear complexity $\mathcal{O}(1\! \times \!N)$. Furthermore, by maintaining a queue-style KV-cache with a fixed capacity $Q$, the computational cost is bounded by $\mathcal{O}(1\! \times \!Q)$ for long-horizon scalability. 

Finally, we investigate several simple strategies of token dropping, including first-in-first-out, random dropping, token merging via interpolation of neighboring tokens, and stride-based dropping. The ablation results in \cref{sec:ablation} justify the final design choice and, more importantly, demonstrate the effectiveness of the queue-style KV caching mechanism.

\begin{figure}[!t]
    \centering
    \includegraphics[width=\linewidth]{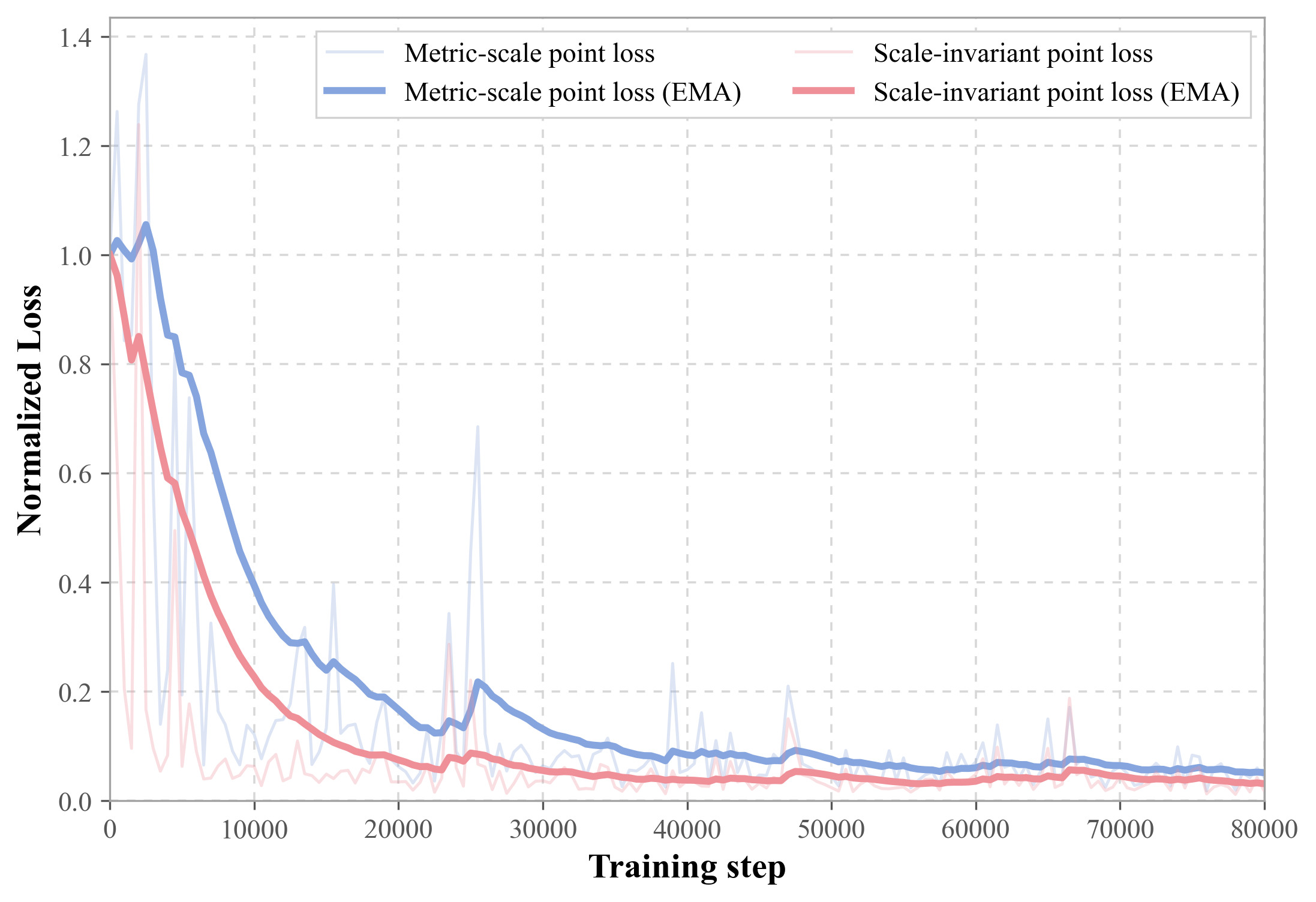}
    \caption{\rev{Training loss curves for the metric-scale and scale-invariant point losses. For clearer comparison, each loss is normalized by its value at the initial training step.}}
    \label{fig:convergence}
\end{figure}

\subsection{Scale-Adaptive Geometry Loss}
\label{sec:metricscale}

\rev{Due to inherent scale ambiguity \cite{ranftl2020midas}, recovering metric scale remains a long-standing challenge in geometry perception \cite{wang2025moge2}. Many existing methods therefore adopt scale-invariant learning objectives \cite{wang2024dust3r}, avoiding the more challenging task of recovering metric scale. However, such objectives discard the physical units of scenes that is essential for real-world geometric understanding.}

\rev{In this work, we propose a scale-adaptive geometry loss that combines the advantages of both scale-invariant and metric-scale learning. This loss implicitly encourages a progressive transition from learning easier scale-invariant geometry to recovering more challenging metric-scale geometry, thereby improving training stability.}

\noindent \textbf{Scale-Adaptive Assumption.} \rev{Inspired by G2-MonoDepth \cite{wang2023g2} in depth estimation, we factorize the metric-scale assumption into scale-invariant constraints and a partial metric-scale term. The former focuses on learning geometric structure; the latter gradually recovers metric scale.}

Under the scale-invariant assumption, the predicted and ground-truth point maps satisfy $\mathbf{\hat{X}}_i / \hat{s} \!=\! \mathbf{X}_i / s$, where $\hat{s}$ and $s$ denote the corresponding global scales. In our framework, each point map $\mathbf{X}_i$ is parameterized by a local point map $\mathbf{P}_i$ predicted by a DPT head \cite{ranftl2021vision} together with the corresponding 12D camera extrinsics $[\mathbf{R}|\mathbf{T}]_i$ predicted by a camera head, where $\mathbf{P}_i\! =\! \mathbf{R}_i \mathbf{X}_i \!+\! \mathbf{T}_i$. Therefore, the scale-invariant constraint can be rewritten as
\begin{equation}
\mathbf{\hat{R}}_i^{-1}(\frac{\mathbf{\hat{P}}_i}{\hat{s}}-\frac{\mathbf{\hat{T}}_i}{\hat{s}})
= \mathbf{R}_i^{-1}(\frac{\mathbf{P}_i}{s}-\frac{\mathbf{T}_i}{s}),
\label{eq:scale-invariant}
\end{equation}
where the sufficient solution corresponds to $\mathbf{\hat{R}}_i \!=\! \mathbf{R}_i$, $\mathbf{\hat{P}}_i/\hat{s} \!=\! \mathbf{P}_i/s$, and $\mathbf{\hat{T}}_i/\hat{s} \!=\! \mathbf{T}_i/s$. 

By additionally imposing a metric-scale constraint $\mathbf{\hat{P}}_i \!=\! \mathbf{P}_i$, the predicted scale $\hat{s}$ is implicitly driven toward the closed-form solution $\hat{s} \!=\! s$. When training has sufficiently converged to a scale-invariant geometry, the relative translation relationship $\mathbf{\hat{T}}_i/\hat{s} \!=\! \mathbf{T}_i/s$ naturally leads to the metric-scale solution $\mathbf{\hat{T}}_i \!=\! \mathbf{T}_i$. The same property also holds for global point maps, leading to the metric-scale consistency $\mathbf{\hat{X}}_i=\mathbf{X}_i$.

This design avoids over-constraining the model with metric-scale supervision. \rev{During training, we observe the intended curriculum behavior, as shown in \cref{fig:convergence}. The scale-invariant loss decreases rapidly at the early stage, whereas the metric-scale loss converges more gradually.} The ablation study in \cref{sec:ablation} further demonstrates the effectiveness of this scale-adaptive design.

\begin{figure}[!t]
    \centering
    \includegraphics[width=0.85\linewidth]{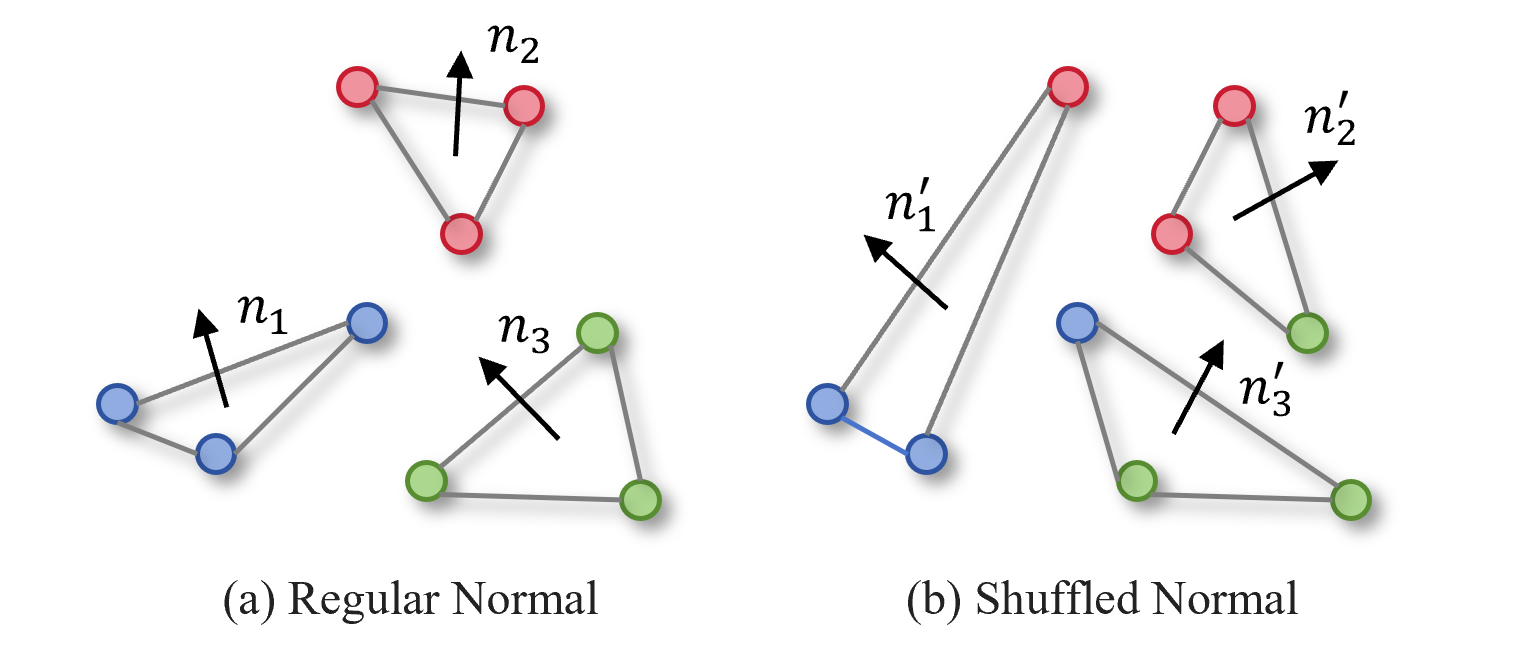}
    \caption{Illustration of two types of normal $n_i$: (a) Regular normal, computed on local surfaces within each frame to enforce local geometric consistency; (b) Shuffled normal, constructed on randomly formed virtual surfaces across frames to encourage global consistency. Points sharing the same color denote pixels originating from the same frame.}
    \label{fig:shufflenormal}
\end{figure}

\noindent \textbf{Scale-Adaptive Geometry Loss.} According to the scale-adaptive assumption, we first build scale-invariant constraints on the local point map and camera extrinsics. Since the scale-invariant camera loss has been defined in \cref{eq:loss_rel_cam}, we impose a scale-invariant constraint \cite{wang2024dust3r} on the predicted local point map $\mathbf{\hat{P}}_i$ and its ground-truth counterpart $\mathbf{P}_i$ for the $i$-th view:
\begin{equation}
\mathcal{L}_{rel}^{point}
= \frac{1}{N} \sum_{i=1}^{N}
\left(\frac{1}{\mathbf{D}_i}\left\| \frac{\mathbf{\hat{P}}_i}{\hat{s}} -\frac{\mathbf{P}_i}{s} \right\|_1 \right),
\label{eq:loss_rel_point}
\end{equation}
where $\hat{s}$ and $s$ are the global scale factors shared with \cref{eq:loss_rel_cam}. The ground-truth depth map $\mathbf{D}_i$ is introduced as a normalization factor to mitigate numerical imbalance across different depth ranges.

For the metric-scale component, we adopt a confidence-aware regression loss on local point maps defined as follows:
\begin{equation}
\mathcal{L}_{abs}^{point}
= \frac{1}{N} \sum_{i=1}^{N}
\left(\frac{\mathbf{C}_i}{\mathbf{D}_i}\left\| \mathbf{\hat{P}}_i -\mathbf{P}_i \right\|_1- \alpha \log \mathbf{C}_i \right),
\label{eq:loss_abs_point}
\end{equation}
where $\mathbf{C}_i$ denotes the predicted confidence map for the $i$-th view. The depth map $\mathbf{D}_i$ serves as balancing factor. The hyperparameter $\alpha$ is fixed to $0.2$.

Beyond these constraints, we introduce a \textit{shuffled normal loss} $\mathcal{L}^{snormal}$ on the global point map predictions $\mathbf{\hat{X}}_i$. By applying the regular normal loss \cite{wang2025vggt} to randomly shuffled pixels across all frames, it enforces global geometric consistency on virtual surfaces \cite{yin2021virtual} across different views, as illustrated in \cref{fig:shufflenormal}~(b). Importantly, the anchor-free camera head in \cref{eq:cam_head} ensures that this shuffled normal loss is also formulated without relying on a fixed reference view.

Finally, the overall training objective is given by
\begin{equation}
\mathcal{L}
= \mathcal{L}_{rel}^{cam} + \mathcal{L}_{rel}^{point}+\mathcal{L}_{abs}^{point}+\mathcal{L}^{snormal}+\mathcal{L}^{normal},
\label{eq:loss_all}
\end{equation}
where $\mathcal{L}^{normal}$ denotes the regular normal loss applied to the local point maps to enforce local geometric consistency, as illustrated in \cref{fig:shufflenormal}~(a).

\subsection{Implementation Details}
\label{sec:implementation}
\noindent \textbf{Training Datasets.} We construct a hybrid training collection by aggregating 21 \textit{metric-scale} datasets. As summarized in \cref{tab:traindata}, these datasets cover a wide range of scenarios, including indoor, outdoor, object-centric, and human-centric settings, while spanning both real-world and synthetic data sources. This diverse composition provides complementary scene geometries, camera types, motion patterns, and scale distributions, enabling the model to learn consistent representations across heterogeneous domains.

\rev{During training, datasets are sampled for each batch according to predefined ratios, which are empirically determined based on the number and diversity of scenes in each dataset. All datasets are publicly available, and many of them have been widely adopted by our baseline methods \cite{wang2025vggt, wang2025continuous, wang2025pi, keetha2025mapanything}. We use only the official training splits and exclude sequences with image resolutions below 518 pixels or lacking metric-scale annotations. These filtering criteria ensure that all training samples provide reliable metric-scale supervision at sufficiently high image resolutions.}

\noindent \rev{\textbf{Modal Tokenizers.} Our modal tokenizers encode auxiliary modalities into point tokens and pose tokens. For point tokens, we first concatenate a local ray map derived from camera intrinsics with the corresponding depth map to construct a ray-depth map. The resulting map is first patchified using the same patch size as the image input, and then projected into point tokens through an MLP. Compared with local point maps, the ray-depth map preserves camera intrinsic information even in regions with zero depth. For pose tokens, we flatten the camera extrinsics into a 12D vector and map it to a single pose token using an MLP. Both tokenizers are applied only once, and the resulting modal tokens are subsequently reused by the modal attention modules across all four stages.}

\noindent \textbf{Multi-Modal Sampling.} Our model supports three optional modalities, including depth maps, camera intrinsics, and camera extrinsics. Following \citet{keetha2025mapanything}, image-only sequences are sampled with a probability of 10\% during training, while mixed multi-modal sequences are sampled with a probability of 90\%. In the multi-modal setting, each modality is independently sampled with a probability of 50\% to simulate diverse modality combinations encountered in real-world scenarios.

To support depth maps from diverse sensors, we adopt the depth pattern simulator from \citet{wang2025pacgdc}, which generates various sampling patterns, including uniform sampling with densities ranging from 0\% to 100\%, LiDAR patterns ranging from 1-beam to 128-beam, SfM feature points extracted using SIFT descriptors, and super-resolution grid patterns with downsampling factors ranging from 1 to 16.

\noindent \rev{\textbf{Camera Head.} We modify the original VGGT camera head in three aspects. First, we remove the original iterative refinement scheme, eliminating three redundant forward passes and thereby reducing the computational cost of the camera head by 75\%. More importantly, this modification substantially simplifies KV-cache management in our framework. Second, we replace the quaternion-based rotation with a 9D representation followed by SVD-based orthogonalization \cite{levinson2020analysis}, which provides a continuous parameterization \cite{zhou2019continuity}. Third, we express all predictions relative to the first, eliminating the shared global transformation.}

\begin{table*}[!t]
    \centering
\ijcvtablelayout{1}

    \caption{Metric-scale training datasets}
    \label{tab:traindata}

    \begin{tabular}{lccccc}
        \toprule
        \textbf{Dataset} & \textbf{Type} & \textbf{Scene} & \textbf{Source} & \textbf{Dynamic} & \textbf{Ratio} \\
        \midrule
        ScanNet++ \cite{yeshwanth2023scannet++}            & Indoor  & 1006  & Real      &        & 8.2\% \\
        ARKitScenes \cite{baruch2021arkitscenes}          & Indoor  & 1661  & Real      &        & 3.0\% \\
        ScanNet \cite{dai2017scannet}             & Indoor  & 1513  & Real      &        & 5.9\% \\
        Matterport3D \cite{chang2017matterport3d}        & Indoor  & 90    & Real      &        & 4.4\% \\
        DynReplica  \cite{karaev2023dynamicstereo}         & Indoor  & 524   & Synthetic & \checkmark & 6.2\% \\
        Hypersim \cite{roberts2021hypersim}            & Indoor  & 461   & Synthetic &        & 5.8\% \\
        Waymo \cite{sun2020scalability}               & Outdoor & 798   & Real      & \checkmark & 7.9\% \\
        Mapfree \cite{arnold2022map}             & Outdoor & 460   & Real      &        & 6.2\% \\
        VKITTI \cite{cabon2020virtual}              & Outdoor & 5     & Synthetic & \checkmark & 1.7\% \\
        MVS-Synth \cite{huang2018deepmvs}            & Outdoor & 120   & Synthetic & \checkmark & 3.6\% \\
        ParaDomain4D  \cite{van2024generative}       & Outdoor & 1528  & Synthetic & \checkmark & 8.0\% \\
        GTA-SfM  \cite{GTA-SFM}       & Outdoor & 200  & Synthetic & \checkmark & 4.9\% \\
        MatrixCity  \cite{li2023matrixcity}       & Outdoor & 24  & Synthetic &  & 3.9\% \\
        Mid-Air  \cite{Fonder2019MidAir}      & Outdoor & 7  & Synthetic &  & 4.9\% \\
        UnrealStereo4K \cite{tosi2021smd}      & Mix     & 9     & Synthetic &        & 1.0\% \\
        TartanAir \cite{wang2020tartanair}           & Mix     & 1037  & Synthetic & \checkmark & 7.3\% \\
        PointOdyssey \cite{zheng2023pointodyssey}        & Mix     & 159   & Synthetic & \checkmark & 3.0\% \\
        Spring \cite{mehl2023spring}              & Mix     & 38    & Synthetic & \checkmark & 0.6\% \\
        WildRGBD  \cite{xia2024rgbd}           & Object & 2754 & Real      &        & 7.0\% \\
        Omniobject3D  \cite{wu2023omniobject3d}       & Object & 6000  & Synthetic &        & 3.3\% \\
        HuMMan  \cite{cai2022humman}       & Human & 907  & real &        \checkmark & 3.0\% \\
        \bottomrule
    \end{tabular}
\end{table*}

\begin{table*}[!t]

\centering
\ijcvtablelayout{2}
\caption{Model complexity}
\label{tab:complexity}

\begin{tabular}{l c c c c c c}
\toprule
\textbf{Method}
& \textbf{Online}
& \textbf{Metric-Scale}
& \textbf{Multi-Modal}
& \textbf{Parameters (B)}
& \textbf{FPS (Img/s)}
& \textbf{Memory (GiB)} \\
\midrule
VGGT\cite{wang2025vggt}        &  &  &  & 1.19 & 31.98 & 11.7 \\
$\pi^3$\cite{wang2025pi}         &  &  &  & 0.96 & 46.18 & 6.4 \\
MapAnything\cite{keetha2025mapanything}      &  & $\checkmark$ & $\checkmark$ & 0.56 & 23.77 & 15.9 \\
DepthAnything3\cite{lin2025depthany3}        &  & $\checkmark$ & $\checkmark$ & 1.40 & 22.36 & 11.4 \\

\rowcolor{gray!15}
\textbf{Ours($G\!=\!N$)} &  & $\checkmark$ & $\checkmark$ & 1.18 & 33.83 & 8.1 \\
\arrayrulecolor{gray}
\midrule
\arrayrulecolor{black}
CUT3R\cite{wang2025continuous}      & $\checkmark$ & $\checkmark$ &  & 0.79 & 19.64 & 4.7 \\
StreamVGGT\cite{zhuo2025streaming} & $\checkmark$ &  &  & 1.19 & 11.50 & 9.6 \\
\revrow TTT3R\cite{chen2025ttt3r} & $\checkmark$ & $\checkmark$  &  & 0.79 & 24.46 & 5.0 \\
\revrow Stream3R\cite{lan2026stream3r} & $\checkmark$ & $\checkmark$  &  & 1.19 & 12.82 & 8.5 \\
\rowcolor{gray!15}
\textbf{Ours($G\!=\!1,Q\!=\!1$)} & $\checkmark$ & $\checkmark$ & $\checkmark$ & 1.18  & 20.41 & 6.7 \\
\rowcolor{gray!15}
\textbf{Ours($G\!=\!1,Q\!=\!N/3$)} & $\checkmark$ & $\checkmark$ & $\checkmark$ & 1.18  & 16.44 & 7.4 \\
\rowcolor{gray!15}
\textbf{Ours($G\!=\!1,Q\!=\!N$)} & $\checkmark$ & $\checkmark$ & $\checkmark$ & 1.18  & 13.38 & 9.1 \\
\bottomrule
\end{tabular}
\par\smallskip
\begin{minipage}{\linewidth}\raggedright
\textit{The results are evaluated in the image-only setting at a resolution of 448$\times$224,}
\textit{using sequences of 50 images on a single 4090D GPU.}
\end{minipage}

\end{table*}

\begin{figure*}[!t]
    \centering
    \includegraphics[width=0.98\linewidth]{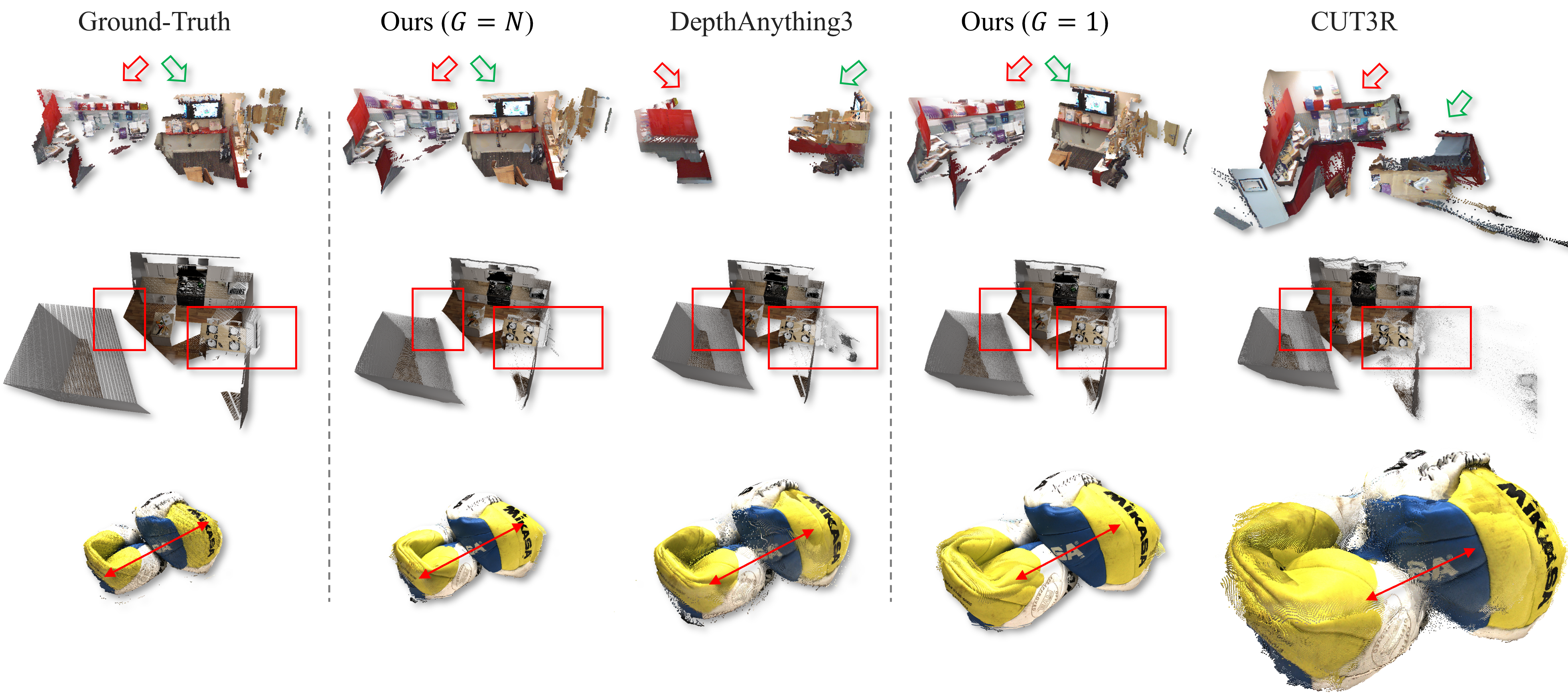}
    \caption{Qualitative results on multi-view reconstruction. All point clouds are presented in their raw form, without any alignment or filtering. Point clouds within the same row are displayed at a consistent scene scale.}
    \label{fig:visualrecon}
\end{figure*}

\noindent \textbf{Training Details.} Our model is initialized with VGGT's pretrained weights. The model is optimized using AdamW with layer-wise learning rates, set to $1\!\times\!10^{-5}$ for pretrained parameters and $1\!\times\!10^{-4}$ for newly introduced ones, while the DINO encoder is kept frozen throughout training. 

Training is performed for 80K iterations at a resolution of 518, with randomly sampled aspect ratios in the range $[0.33, 1.0]$, and a dynamic sequence length ranging from 12 to 24. To accommodate different view configurations, the group size $G$ is randomly sampled from 1 to 24 during training. All other training settings follow VGGT, including 5\% warm-up schedule, cosine learning rate decay, gradient norm clipping, bfloat16 precision, and gradient checkpointing. Training is performed on 64 H100 GPUs with 48 images per GPU and takes over 7 days. 

In addition, we remove other heads originally used in VGGT to reduce unnecessary computational overhead.

\section{Experiments}
\label{sec:exp}

\subsection{Experiment Setting}

\rev{To evaluate \textsc{UniT} as a unified geometry learning formulation, we conduct comprehensive experiments on eleven benchmark datasets spanning seven representative geometry perception tasks and compare against nine strong baselines.}

\noindent \textbf{Tasks and Datasets.} We select seven representative tasks to comprehensively evaluate our model under diverse settings. The evaluation protocol follows prior work \cite{wang2025continuous,wang2025pi,deng2025vggt}, while the dataset setup is slightly revised so that all datasets can be evaluated under both metric-scale and multi-modal settings.
\begin{itemize}
    \item \textit{Multi-view reconstruction} is evaluated on the scene-level real-world 7-Scenes \cite{shotton2013scene} and synthetic NRGBD \cite{azinovic2022neural} datasets, as well as the object-centric DTU \cite{jensen2014large} dataset; 
    \item \textit{Camera pose estimation} is conducted on the synthetic outdoor Sintel\cite{butler2012naturalistic} dataset and the real-world indoor TUM-Dynamic \cite{sturm2012benchmark} dataset and \rev{ScanNetv2 \cite{dai2017scannet} test set}; 
    \item \textit{Video depth estimation} is evaluated on Sintel and the real-world Bonn\cite{palazzolo2019refusion} and ETH3D\cite{schops2017multi} datasets; 
    \item \textit{Monocular depth estimation} is assessed on Sintel, and the widely used KITTI\cite{geiger2013vision} and NYUv2\cite{silberman2012indoor} datasets; 
    \item \rev{\textit{Long-horizon perception} adopts indoor NRGBD dataset using 500-frame sequences and on the outdoor KITTI-Odometry\cite{geiger2012we} dataset using kilometer-scale trajectories;}
    \item \textit{Multi-modal reconstruction} includes arbitrary combinations \cite{keetha2025mapanything} of depth maps, camera intrinsics, and extrinsics on 7-Scenes, ETH3D, and ScanNetv2;
    \item \textit{Depth completion} evaluates depth maps with four sparse patterns \cite{wang2025pacgdc} on Sintel, KITTI, and NYUv2 datasets.
\end{itemize}

\noindent \textbf{Metrics.} Point maps are evaluated using Accuracy (Acc.), Completion (Comp.), and Normal Consistency (N.C.). Camera poses are assessed using Absolute Trajectory Error (ATE), Relative Pose Error for translation (RPE\textsubscript{tra}), and Relative Pose Error for rotation (RPE\textsubscript{rot}). For depth maps, we report Absolute Relative Error (AbsRel), Root Mean Square Error (RMSE), and prediction accuracy under the threshold $\delta\! <\! 1.25$. Moreover, the average rank (Rank)\cite{wang2023g2,lin2025depthany3} is reported to summarize overall performance.

All metrics are reported in meters (m) under two scale settings. In the scale-invariant setting, $\mathrm{Sim}(3)$ or median alignment is applied to resolve scale ambiguity. In the metric-scale setting, scale adjustment is disabled.

\begin{table*}[t]
\centering
\ijcvtablelayout{3}

\caption{Multi-view reconstruction on 7-Scenes, NRGBD, and DTU under different alignment settings.}
\label{tab:pointmap}

\begin{tabular}{c l c c c c c c c c c c c}
\toprule
\multirow{2}{*}{\textbf{Scale}}
& \multirow{2}{*}{\textbf{Method}} 
& \multirow{2}{*}{\textbf{Online}} 
& \multicolumn{3}{c}{\textbf{7-Scenes}} 
& \multicolumn{3}{c}{\textbf{NRGBD}} 
& \multicolumn{3}{c}{\textbf{DTU}} 
& \multirow{2}{*}{\textbf{Rank$\downarrow$}} \\

\cmidrule(lr){4-6}
\cmidrule(lr){7-9}
\cmidrule(lr){10-12}

& & 
& Acc.$\downarrow$ & Comp.$\downarrow$ & N.C.$\uparrow$
& Acc.$\downarrow$ & Comp.$\downarrow$ & N.C.$\uparrow$
& Acc.$\downarrow$ & Comp.$\downarrow$ & N.C.$\uparrow$
& \\
\midrule

\multirow{10}{*}{\shortstack{\textit{Scale-} \\ \textit{Invariant}}}       
& VGGT\cite{wang2025vggt} &  & \underline{0.043} & \underline{0.056} & 0.732 & 0.052 & 0.068 & 0.884 & \underline{0.001} & 0.002 & 0.666 & 3.33 \\
& $\pi^3$\cite{wang2025pi} &        & 0.047 & 0.073 & \underline{0.742} & \textbf{0.024} & \underline{0.029} & \underline{0.908} & \textbf{0.001} & \underline{0.002} & 0.668 & \textbf{2.11} \\
& MapAnything\cite{keetha2025mapanything} &        & 0.070 & 0.089 & 0.727 & 0.089 & 0.093 & 0.837 & 0.002 & 0.002 & \underline{0.680} & 4.33 \\
& DepthAnything3\cite{lin2025depthany3} &        & 0.054 & 0.101 & 0.734 & \underline{0.030} & \textbf{0.029} & \textbf{0.916} & 0.002 & \textbf{0.002} & 0.677 & 2.67 \\
& \cellcolor{gray!15}\textbf{Ours($G\!=\!N$)} & \cellcolor{gray!15}        & \cellcolor{gray!15} \textbf{0.027}    & \cellcolor{gray!15} \textbf{0.032}    & \cellcolor{gray!15} \textbf{0.782}    & \cellcolor{gray!15} 0.040    & \cellcolor{gray!15} 0.045    & \cellcolor{gray!15} 0.904    & \cellcolor{gray!15} 0.003    & \cellcolor{gray!15} 0.002    & \cellcolor{gray!15} \textbf{0.683}    & \cellcolor{gray!15} \underline{2.56}    \\
\arrayrulecolor{gray}
\cmidrule(lr){2-13}
\arrayrulecolor{black}
& CUT3R\cite{wang2025continuous}  &  $\checkmark$     & \underline{0.087} & 0.095 & 0.707 & 0.104 & 0.078 & 0.823 & \underline{0.005} & 0.003 & 0.681 & 3.44 \\
& StreamVGGT\cite{zhuo2025streaming}  &  $\checkmark$      & 0.100 & \underline{0.091} & \underline{0.723} & \underline{0.073} & \underline{0.063} & \underline{0.874} & \textbf{0.002} & \textbf{0.002} & \underline{0.681} & \underline{2.11} \\
& \revrow TTT3R\cite{chen2025ttt3r}  &  $\checkmark$  & 0.094 & 0.134 & 0.682 & 0.104 & 0.079 & 0.818 & 0.005 & 0.004 & 0.679 & 4.56 \\ 
& \revrow Stream3R\cite{lan2026stream3r}  &  $\checkmark$      & 0.082 & 0.113 & 0.715 & 0.100 & 0.088 & 0.854 & 0.003 & \underline{0.002} & 0.678 & 3.22 \\
& \cellcolor{gray!15}\textbf{Ours($G\!=\!1$)}        & \cellcolor{gray!15}$\checkmark$         & \cellcolor{gray!15} \textbf{0.060}    & \cellcolor{gray!15} \textbf{0.063}    & \cellcolor{gray!15} \textbf{0.748}    & \cellcolor{gray!15} \textbf{0.042}    & \cellcolor{gray!15} \textbf{0.042}    & \cellcolor{gray!15} \textbf{0.897}    & \cellcolor{gray!15} 0.006    & \cellcolor{gray!15} 0.003    & \cellcolor{gray!15} \textbf{0.683}    & \cellcolor{gray!15} \textbf{1.67}    \\
\midrule

\multirow{7}{*}{\shortstack{\textit{Metric-} \\ \textit{Scale}}} 
& MapAnything\cite{keetha2025mapanything}  &  & 0.406 & 0.166 & 0.588 & 0.206 & 0.164 & 0.765 & 1.146 & 0.064 & 0.570 & 3.00 \\
& DepthAnything3\cite{lin2025depthany3}   &        & \underline{0.074} & \underline{0.087} & \underline{0.720} & \textbf{0.121} & \textbf{0.157} & \textbf{0.802} & \underline{0.664} & \underline{0.035} & \underline{0.597} & \underline{1.67} \\
& \cellcolor{gray!15}\textbf{Ours($G\!=\!N$)}  & \cellcolor{gray!15} &  \cellcolor{gray!15} \textbf{0.047} & \cellcolor{gray!15} \textbf{0.042} & \cellcolor{gray!15} \textbf{0.751} & \cellcolor{gray!15} \underline{0.158} & \cellcolor{gray!15} \underline{0.159} & \cellcolor{gray!15} \underline{0.792} & \cellcolor{gray!15} \textbf{0.055} & \cellcolor{gray!15} \textbf{0.007} & \cellcolor{gray!15} \textbf{0.646} & \cellcolor{gray!15} \textbf{1.33} \\
\arrayrulecolor{gray}
\cmidrule(lr){2-13}
\arrayrulecolor{black}
& CUT3R\cite{wang2025continuous}       & $\checkmark$       & \underline{0.095} & \underline{0.102} & \underline{0.696} & 0.189 & \underline{0.150} & \underline{0.743} & \underline{0.137} & 0.011 & 0.581 & \underline{2.56} \\
& \revrow TTT3R\cite{chen2025ttt3r}  &  $\checkmark$  & 0.127 & 0.125 & 0.673 & 0.188 & \textbf{0.149} & 0.742 & 0.138 & 0.011 & 0.579 & 3.00 \\ 
& \revrow Stream3R\cite{lan2026stream3r}  &  $\checkmark$      & 0.118 & 0.571 & 0.592 & \underline{0.185} & 1.041 & 0.587 & \textbf{0.023} & \textbf{0.006} & \underline{0.626} & 2.78 \\
& \cellcolor{gray!15}\textbf{Ours($G\!=\!1$)} & \cellcolor{gray!15}$\checkmark$        & \cellcolor{gray!15} \textbf{0.081}    & \cellcolor{gray!15} \textbf{0.062}    & \cellcolor{gray!15} \textbf{0.723}    & \cellcolor{gray!15} \textbf{0.157}    & \cellcolor{gray!15} 0.158    & \cellcolor{gray!15} \textbf{0.790}    & \cellcolor{gray!15} 0.152    & \cellcolor{gray!15} \underline{0.007}    & \cellcolor{gray!15} \textbf{0.647}    & \cellcolor{gray!15} \textbf{1.67}    \\

\bottomrule
\end{tabular}

\end{table*}

\begin{table*}[t]
\centering
\ijcvtablelayout{4}
\caption{Camera pose estimation on Sintel, TUM-Dynamic, and ScanNetv2 under different alignment settings.}
\label{tab:pose}

\begin{tabular}{c l c c c c c c c c c c c}
\toprule
\multirow{2}{*}{\textbf{Scale}}
& \multirow{2}{*}{\textbf{Methods}} 
& \multirow{2}{*}{\textbf{Online}} 
& \multicolumn{3}{c}{\textbf{Sintel}} 
& \multicolumn{3}{c}{\textbf{TUM-Dynamic}} 
& \multicolumn{3}{c}{\textbf{ScanNetv2}} 
& \multirow{2}{*}{\textbf{Rank$\downarrow$}} \\

\cmidrule(lr){4-6}
\cmidrule(lr){7-9}
\cmidrule(lr){10-12}

& & 
& ATE $\downarrow$ & RPE\textsubscript{tra}$\downarrow$ & RPE\textsubscript{rot}$\downarrow$
& ATE $\downarrow$ & RPE\textsubscript{tra}$\downarrow$ & RPE\textsubscript{rot}$\downarrow$
& ATE $\downarrow$ & RPE\textsubscript{tra}$\downarrow$ & RPE\textsubscript{rot}$\downarrow$
& \\
\midrule

\multirow{10}{*}{\shortstack{\textit{Scale-} \\ \textit{Invariant}}}       
& VGGT\cite{wang2025vggt} &  & 0.169 & 0.062 & 0.476 & \textbf{0.012} & 0.010 & 0.325 & 0.032 & 0.015 & 0.382 & 3.33 \\
& $\pi^3$\cite{wang2025pi} &        & \textbf{0.073} & \textbf{0.038} & \textbf{0.288} & 0.015 & \underline{0.009} & \underline{0.325} & 0.032 & \textbf{0.013} & \textbf{0.360} & \textbf{1.56} \\
& MapAnything\cite{keetha2025mapanything} &        & 0.215 & 0.082 & 3.091 & 0.029 & 0.018 & 0.434 & 0.093 & 0.032 & 1.158 & 5.00 \\
& DepthAnything3\cite{lin2025depthany3} &        & \underline{0.108} & \underline{0.048} & \underline{0.366} & \underline{0.012} & \textbf{0.009} & 0.325 & 0.033 & 0.014 & 0.468 & 2.56 \\
& \cellcolor{gray!15}\textbf{Ours($G\!=\!N$)} &  \cellcolor{gray!15} & \cellcolor{gray!15}0.124 & \cellcolor{gray!15}0.059 & \cellcolor{gray!15}0.455 & \cellcolor{gray!15}0.013 & \cellcolor{gray!15}0.010 & \cellcolor{gray!15}\textbf{0.325} & \cellcolor{gray!15}\textbf{0.031} & \cellcolor{gray!15}\underline{0.014} & \cellcolor{gray!15}\underline{0.380} & \cellcolor{gray!15}\underline{2.56} \\
\arrayrulecolor{gray}
\cmidrule(lr){2-13}
\arrayrulecolor{black}
& CUT3R\cite{wang2025continuous}  &  $\checkmark$     & \underline{0.227} & \textbf{0.065} & \textbf{0.635} & 0.044 & 0.015 & 0.461 & 0.096 & 0.022 & 0.549 & 4.00 \\
& StreamVGGT\cite{zhuo2025streaming}  &  $\checkmark$      & 0.227 & 0.099 & 0.847 & \underline{0.027} & \textbf{0.012} & \textbf{0.328} & \underline{0.045} & \underline{0.017} & \underline{0.467} & \underline{2.78} \\
& \revrow TTT3R\cite{chen2025ttt3r}  &  $\checkmark$  & 0.206 & 0.086 & \underline{0.690} & 0.030 & 0.014 & 0.404 & 0.065 & 0.020 & 0.534 & 3.56 \\ 
& \revrow Stream3R\cite{lan2026stream3r}  &  $\checkmark$      & 0.219 & 0.072 & 0.901 & 0.027 & 0.013 & \underline{0.342} & 0.045 & 0.019 & 0.517 & 3.00 \\
& \cellcolor{gray!15}\textbf{Ours($G\!=\!1$)}        & \cellcolor{gray!15}$\checkmark$         & \cellcolor{gray!15}\textbf{0.143}    & \cellcolor{gray!15}\underline{0.067}    & \cellcolor{gray!15}0.766    & \cellcolor{gray!15}\textbf{0.024}    & \cellcolor{gray!15}\underline{0.013}    & \cellcolor{gray!15}0.344    & \cellcolor{gray!15}\textbf{0.037}    & \cellcolor{gray!15}\textbf{0.016}    & \cellcolor{gray!15}\textbf{0.424}    & \cellcolor{gray!15}\textbf{1.67}    \\
\midrule

\multirow{7}{*}{\shortstack{\textit{Metric-} \\ \textit{Scale}}} & MapAnything\cite{keetha2025mapanything}  &  & 0.479 & 0.148 & 2.743 & 0.072 & 0.025 & 0.430 & 0.148 & 0.040 & 1.155 & 3.00 \\
& DepthAnything3\cite{lin2025depthany3}   &        & \underline{0.394} & \textbf{0.036} & \textbf{0.366} & 0.019 & \textbf{0.009} & \underline{0.325} & \underline{0.044} & \underline{0.014} & \underline{0.468} & \underline{1.56} \\
& \cellcolor{gray!15}\textbf{Ours($G\!=\!N$)}  &  \cellcolor{gray!15}  & \cellcolor{gray!15}\textbf{0.331} & \cellcolor{gray!15}\underline{0.052} & \cellcolor{gray!15}\underline{0.455} & \cellcolor{gray!15}\textbf{0.017} & \cellcolor{gray!15}\underline{0.010} & \cellcolor{gray!15}\textbf{0.325} & \cellcolor{gray!15}\textbf{0.038} & \cellcolor{gray!15}\textbf{0.014} & \cellcolor{gray!15}\textbf{0.380} & \cellcolor{gray!15}\textbf{1.44} \\
\arrayrulecolor{gray}
\cmidrule(lr){2-13}
\arrayrulecolor{black}
& CUT3R\cite{wang2025continuous}       & $\checkmark$       & \underline{0.424} & \underline{0.053} & \textbf{0.635} & 0.060 & 0.018 & 0.461 & 0.112 & 0.021 & 0.549 & 2.89 \\
& \revrow TTT3R\cite{chen2025ttt3r}  &  $\checkmark$  & \textbf{0.412} & 0.058 & \underline{0.690} & \underline{0.042} & 0.016 & 0.404 & \underline{0.080} & \underline{0.019} & 0.534 & \underline{2.33} \\ 
& \revrow Stream3R\cite{lan2026stream3r}  &  $\checkmark$      & 0.598 & \textbf{0.050} & 0.901 & 0.181 & \underline{0.013} & \textbf{0.342} & 0.244 & 0.022 & \underline{0.517} & 2.89 \\
& \cellcolor{gray!15}\textbf{Ours($G\!=\!1$)} & \cellcolor{gray!15}$\checkmark$        & \cellcolor{gray!15}0.480    & \cellcolor{gray!15}0.100    & \cellcolor{gray!15}0.766    & \cellcolor{gray!15}\textbf{0.034}    & \cellcolor{gray!15}\textbf{0.013}    & \cellcolor{gray!15}\underline{0.344}    & \cellcolor{gray!15}\textbf{0.045}    & \cellcolor{gray!15}\textbf{0.016}    & \cellcolor{gray!15}\textbf{0.424}    & \cellcolor{gray!15}\textbf{1.89}    \\

\bottomrule
\end{tabular}

\end{table*}

\noindent \textbf{Baselines.} We select nine representative methods that differ in view configurations, scale assumptions, and multi-modal settings. As shown in \cref{tab:complexity}, VGGT\cite{wang2025vggt} and $\pi^3$\cite{wang2025pi} are offline models operating under scale-invariant settings. MapAnything\cite{keetha2025mapanything} and DepthAnything3 (Nested)\cite{lin2025depthany3} are offline models that perform metric-scale estimation with multi-modal integration. \rev{CUT3R\cite{wang2025continuous} and TTT3R\cite{chen2025ttt3r} are RNN-style online models with metric-scale inference, while StreamVGGT\cite{zhuo2025streaming} and Stream3R\cite{lan2026stream3r} adopt online inference under scale-invariant and metric-scale settings, respectively. VGGT-Long\cite{deng2025vggt} serves as a long-horizon baseline using an optimization-based strategy.} Together, these methods cover a broad range of design choices and enable comprehensive comparisons across different settings. Notably, since DepthAnything3 only supports camera parameter integration, we adopt scale alignment to ensure comparability when incorporating other modalities.

In addition, we evaluate the computational complexity of these baselines in \cref{tab:complexity}, including the number of parameters, frames per second (FPS), and maximum GPU memory consumption. The results are measured at a resolution of $448\!\times\!224$ with a sequence length of 50 on a single 4090D GPU. Under the online setting, we report results of our method with three KV-cache queue capacities, namely $Q\!=\!1$, $Q\!=\!N/3$, and $Q\!=\!N$. The results show that our model achieves competitive computational efficiency compared with the baselines.

\noindent \textbf{Organization.} The experimental section is organized into eight subsections, including multi-view reconstruction in \cref{sec:pointmap}, camera pose estimation in \cref{sec:pose}, video depth estimation in \cref{sec:videodepth}, monocular depth estimation in \cref{sec:monoculardepth}, long-horizon perception in \cref{sec:longhorizon}, multi-modal reconstruction in \cref{sec:multimodal}, depth completion in \cref{sec:depthcompletion}, and ablation study in \cref{sec:ablation}.

\begin{table*}[t]
\centering
\ijcvtablelayout{5}
\caption{Video Depth Estimation on Sintel, Bonn, and ETH3D under different alignment settings.}
\label{tab:videodepth}

\begin{tabular}{c l c c c c c c c c c c c}
\toprule
\multirow{2}{*}{\textbf{Scale}}
& \multirow{2}{*}{\textbf{Methods}} 
& \multirow{2}{*}{\textbf{Online}} 
& \multicolumn{3}{c}{\textbf{Sintel}} 
& \multicolumn{3}{c}{\textbf{Bonn}} 
& \multicolumn{3}{c}{\textbf{ETH3D}} 
& \multirow{2}{*}{\textbf{Rank$\downarrow$}} \\

\cmidrule(lr){4-6}
\cmidrule(lr){7-9}
\cmidrule(lr){10-12}

& & 
& AbsRel$\downarrow$ & RMSE$\downarrow$ & $\delta\!<\!1.25$$\uparrow$
& AbsRel$\downarrow$ & RMSE$\downarrow$ & $\delta\!<\!1.25$$\uparrow$
& AbsRel$\downarrow$ & RMSE$\downarrow$ & $\delta\!<\!1.25$$\uparrow$
& \\
\midrule

\multirow{10}{*}{\shortstack{\textit{Scale-} \\ \textit{Invariant}}}       
& VGGT\cite{wang2025vggt} &  & 0.231 & 5.517 & 0.657 & 0.056 & 0.254 & 0.970 & 0.045 & 0.544 & 0.979 & 3.56 \\
& $\pi^3$\cite{wang2025pi} &        & 0.217 & \textbf{3.911} & \textbf{0.714} &  \underline{0.053} &  \underline{0.236} &  \underline{0.972} & \textbf{0.028} & \textbf{0.337} & \textbf{0.998} & \textbf{1.44} \\
& MapAnything\cite{keetha2025mapanything} &        & 0.463 & 5.730 & 0.565 & 0.082 & 0.270 & 0.885 & 0.093 & 0.804 & 0.891 & 5.00 \\
& DepthAnything3\cite{lin2025depthany3} &        & \underline{0.216} &  \underline{4.636} &  \underline{0.700} & \textbf{0.052} & \textbf{0.233} & \textbf{0.972} &  \underline{0.030} & 0.525 &  \underline{0.994} &  \underline{1.89} \\
& \cellcolor{gray!15}\textbf{Ours($G\!=\!N$)} &  \cellcolor{gray!15} & \cellcolor{gray!15}\textbf{0.215} & \cellcolor{gray!15}4.780 & \cellcolor{gray!15}0.654 & \cellcolor{gray!15}0.068 & \cellcolor{gray!15}0.256 & \cellcolor{gray!15}0.965 & \cellcolor{gray!15}0.037 & \cellcolor{gray!15} \underline{0.487} & \cellcolor{gray!15}0.988 & \cellcolor{gray!15}3.11 \\
\arrayrulecolor{gray}
\cmidrule(lr){2-13}
\arrayrulecolor{black}
& CUT3R\cite{wang2025continuous}  &  $\checkmark$     & 0.562 & 7.285 & 0.542 & 0.083 & 0.293 & 0.949 &  0.133 &  1.243 &  0.806 & 4.67 \\
& StreamVGGT\cite{zhuo2025streaming}  &  $\checkmark$      &  0.270 &  5.511 &  0.597 & \textbf{0.060} &  \underline{0.262} & \textbf{0.974} & 0.156 & 1.359 & 0.771 & 3.11 \\
& \revrow TTT3R\cite{chen2025ttt3r}  &  $\checkmark$  & 0.527 & 7.158 & 0.554 & 0.072 & 0.268 & 0.964 & 0.127 & 1.235 & 0.824 & 3.44 \\ 
& \revrow Stream3R\cite{lan2026stream3r}  &  $\checkmark$      & \textbf{0.238} & \textbf{4.178} & \textbf{0.654} & \underline{0.068} & 0.266 & \underline{0.956} & \textbf{0.065} & \textbf{0.677} & \textbf{0.960} & \textbf{1.56} \\
& \cellcolor{gray!15}\textbf{Ours($G\!=\!1$)}        & \cellcolor{gray!15}$\checkmark$         & \cellcolor{gray!15}\underline{0.253}    & \cellcolor{gray!15}\underline{5.130}    & \cellcolor{gray!15}\underline{0.609}    & \cellcolor{gray!15} 0.069    & \cellcolor{gray!15}\textbf{0.260}    & \cellcolor{gray!15} 0.951    & \cellcolor{gray!15}\underline{0.084}    & \cellcolor{gray!15}\underline{0.804}    & \cellcolor{gray!15}\underline{0.944}    & \cellcolor{gray!15}\underline{2.22}    \\
\midrule

\multirow{7}{*}{\shortstack{\textit{Metric-} \\ \textit{Scale}}} 
& MapAnything\cite{keetha2025mapanything}  &  & 1.021 &  \underline{7.064} & \textbf{0.263} & 0.319 & 0.981 & 0.251 & 0.179 & \underline{1.164} & 0.746 & 2.56 \\
& DepthAnything3\cite{lin2025depthany3}   &        & \textbf{0.492} & 7.738 & 0.178 &  \underline{0.147} &  \underline{0.416} &  \underline{0.778} &  \underline{0.127} & \textbf{1.057} &  \underline{0.772} & \underline{2.00} \\
& \cellcolor{gray!15}\textbf{Ours($G\!=\!N$)}  &  \cellcolor{gray!15}  & \cellcolor{gray!15} \underline{0.687} & \cellcolor{gray!15}\textbf{6.763} & \cellcolor{gray!15} \underline{0.258} & \cellcolor{gray!15}\textbf{0.106} & \cellcolor{gray!15}\textbf{0.319} & \cellcolor{gray!15}\textbf{0.892} & \cellcolor{gray!15}\textbf{0.125} & \cellcolor{gray!15}1.434 & \cellcolor{gray!15}\textbf{0.834} & \cellcolor{gray!15}\textbf{1.44} \\
\arrayrulecolor{gray}
\cmidrule(lr){2-13}
\arrayrulecolor{black}
& CUT3R\cite{wang2025continuous}       & $\checkmark$       & 0.837 & 8.703 & \textbf{0.269} & 0.103 & 0.349 & 0.886 & \underline{0.315} & \underline{2.762} & \underline{0.345} & 2.44 \\
& \revrow TTT3R\cite{chen2025ttt3r}  &  $\checkmark$  & 0.841 & \underline{8.671} & \underline{0.262} & \textbf{0.091} & \textbf{0.314} & \textbf{0.942} & 0.333 & 2.902 & 0.286 & \underline{2.22} \\ 
& \revrow Stream3R\cite{lan2026stream3r}  &  $\checkmark$      & \underline{0.760} & 9.761 & 0.033 & 0.681 & 1.870 & 0.004 & 0.797 & 6.009 & 0.000 & 3.67 \\
& \cellcolor{gray!15}\textbf{Ours($G\!=\!1$)} & \cellcolor{gray!15}$\checkmark$        & \cellcolor{gray!15}\textbf{0.776}    & \cellcolor{gray!15}\textbf{7.244}    & \cellcolor{gray!15}0.206    & \cellcolor{gray!15}\underline{0.098}    & \cellcolor{gray!15}\underline{0.318}    & \cellcolor{gray!15}\underline{0.923}    & \cellcolor{gray!15}\textbf{0.177}    & \cellcolor{gray!15}\textbf{1.661}    & \cellcolor{gray!15}\textbf{0.712}    & \cellcolor{gray!15}\textbf{1.67}    \\

\bottomrule
\end{tabular}

\end{table*}

\begin{table*}[t]
\centering
\ijcvtablelayout{6}
\caption{Monocular depth estimation on Sintel, KITTI, and NYUv2 under different alignment settings.}
\label{tab:monocular}

\begin{tabular}{c l c c c c c c c c c c}
\toprule
\multirow{2}{*}{\textbf{Scale}} 
& \multirow{2}{*}{\textbf{Methods}}
& \multicolumn{3}{c}{\textbf{Sintel}} 
& \multicolumn{3}{c}{\textbf{KITTI}} 
& \multicolumn{3}{c}{\textbf{NYUv2}} 
& \multirow{2}{*}{\textbf{Rank$\downarrow$}} \\

\cmidrule(lr){3-5}
\cmidrule(lr){6-8}
\cmidrule(lr){9-11}

& 
& AbsRel$\downarrow$ & RMSE$\downarrow$ & $\delta\!<\!1.25$$\uparrow$
& AbsRel$\downarrow$ & RMSE$\downarrow$ & $\delta\!<\!1.25$$\uparrow$
& AbsRel$\downarrow$ & RMSE$\downarrow$ & $\delta\!<\!1.25$$\uparrow$
& \\
\midrule

\multirow{9}{*}{\shortstack{\textit{Scale-} \\ \textit{Invariant}}} & VGGT \cite{wang2025vggt}            & 0.320 & 5.810 & 0.620 & 0.077 & 4.217 & 0.931 & 0.060 & 0.308 & 0.950 & 5.78 \\
& $\pi^3$\cite{wang2025pi}          & 0.369 & 4.917 & \underline{0.649} & \textbf{0.060} & \textbf{3.182} & \textbf{0.971} & 0.059 & 0.304 & 0.953 & \underline{2.67} \\
& MapAnything\cite{keetha2025mapanything}      & 0.395 & 5.356 & 0.596 & 0.096 & 4.062 & 0.932 & 0.077 & 0.371 & 0.933 & 6.33 \\
& DepthAnything3\cite{lin2025depthany3}   & 0.308 & 5.143 & 0.616 & 0.074 & \underline{3.250} & 0.956 & 0.097 & 0.545 & 0.894 & 5.44 \\
& CUT3R\cite{wang2025continuous}            & 0.480 & 6.116 & 0.524 & 0.103 & 5.308 & 0.885 & 0.092 & 0.449 & 0.897 & 7.67 \\
& StreamVGGT\cite{zhuo2025streaming}       & 0.294 & 5.315 & 0.637 & 0.077 & 3.959 & 0.940 & \textbf{0.055} & \textbf{0.276} & \textbf{0.959} & 3.22 \\
& \revrow TTT3R\cite{chen2025ttt3r}  & 0.480 & 6.116 & 0.524 & 0.103 & 5.308  & 0.885 & 0.092 & 0.449 & 0.897 & 7.67 \\ 
& \revrow Stream3R\cite{lan2026stream3r}   & \textbf{0.251} & \textbf{4.349} & \textbf{0.661} & 0.069 & 3.705  & 0.948 & \underline{0.056} & \underline{0.294} & \underline{0.958} & \textbf{2.22} \\
& \cellcolor{gray!15}\textbf{Ours}    & \cellcolor{gray!15}\underline{0.282}    & \cellcolor{gray!15}\underline{4.732}    & \cellcolor{gray!15}0.642    & \cellcolor{gray!15}\underline{0.061}    & \cellcolor{gray!15}3.976    & \cellcolor{gray!15}\underline{0.966}    & \cellcolor{gray!15}0.058    & \cellcolor{gray!15}0.311    & \cellcolor{gray!15}0.953    & \cellcolor{gray!15}3.00    \\
\midrule

\multirow{6}{*}{\shortstack{\textit{Metric-} \\ \textit{Scale}}} & MapAnything\cite{keetha2025mapanything} & 0.984 & \underline{6.840} & \textbf{0.270} & \textbf{0.132} & \textbf{4.243} & \textbf{0.900} & 0.156 & 0.550 & 0.741 & \underline{3.00} \\
& DepthAnything3\cite{lin2025depthany3}  & \textbf{0.561} & 7.948 & 0.138 & 0.264 & 5.048 & 0.457 & \underline{0.121} & 0.589 & 0.876 & 3.44 \\
& CUT3R\cite{wang2025continuous}           & 0.980 & 8.354 & \underline{0.240} & \underline{0.134} & 5.458 & 0.829 & 0.122 & \underline{0.502} & 0.865 & 3.11 \\
& \revrow TTT3R\cite{chen2025ttt3r}   & 0.980 & 8.354 & 0.240 & 0.134 & 5.458  & 0.829 & 0.122 & 0.502 & 0.865 & 3.11 \\ 
& \revrow Stream3R\cite{lan2026stream3r}  & 0.767 & 9.745 & 0.053 & 0.957 & 18.160 & 0.000 & 0.655 & 2.005 & 0.001 & 5.67 \\
& \cellcolor{gray!15}\textbf{Ours}   & \cellcolor{gray!15}\underline{0.716}    & \cellcolor{gray!15}\textbf{6.615}    & \cellcolor{gray!15}0.200    & \cellcolor{gray!15}0.149    & \cellcolor{gray!15}\underline{4.890}    & \cellcolor{gray!15}\underline{0.833}    & \cellcolor{gray!15}\textbf{0.096}    & \cellcolor{gray!15}\textbf{0.391}    & \cellcolor{gray!15}\textbf{0.921}    & \cellcolor{gray!15}\textbf{1.67}    \\

\bottomrule
\end{tabular}

\end{table*}

\subsection{Multi-View Reconstruction}
\label{sec:pointmap}

Following prior works\cite{wang2025continuous,wang2025pi}, the evaluated frames are sampled with strides of 200, 500, and 5 on 7-Scenes, NRGBD, and DTU datasets, respectively. 

In \cref{tab:pointmap}, all methods are first evaluated in the scale-invariant setting, while methods capable of metric-scale inference are further evaluated in the metric-scale setting. \textsc{UniT} ranks first in the scale-invariant online, metric-scale online, and metric-scale offline settings, and second in the scale-invariant offline setting. These results highlight that, even with a single unified model, \textsc{UniT} achieves strong competitiveness against existing 3D foundation models.

In addition, we present qualitative results in \cref{fig:visualrecon}. The point clouds of the same scene are shown at a consistent scale. This allows a direct comparison of metric scale through their relative sizes in visualization. The results consistently show that \textsc{UniT} recovers more accurate metric-scale geometry while preserving fine-grained structures.

\subsection{Camera Pose Estimation}
\label{sec:pose}
Following prior work \cite{zhang2024monst3r,wang2025continuous,wang2025pi}, we evaluate all frames on the Sintel dataset, and 90 frames per scene on the TUM-Dynamic dataset and the \rev{ScanNetV2 test set}.

In \cref{tab:pose}, we observe a similar trend to that in multi-view reconstruction. \textsc{UniT} achieves the best performance in the scale-invariant online, metric-scale online, and metric-scale offline settings, and ranks second in the scale-invariant offline setting. These results suggest that our model effectively captures metric-scale trajectories while remaining competitive in modeling relative pose relationships.

\subsection{Video Depth Estimation}
\label{sec:videodepth}
Following prior work \cite{zhang2024monst3r,wang2025continuous,wang2025pi}, the evaluated sequences include all frames on Sintel, 110 frames per scene on Bonn, and frames sampled with a stride of 5 on ETH3D.

\cref{tab:videodepth} shows that \textsc{UniT} achieves the best performance in metric-scale settings, ranks second in the scale-invariant online setting, and remains competitive in the scale-invariant offline setting. These results further demonstrate the effectiveness of \textsc{UniT} for dense depth prediction.

\subsection{Monocular Depth Estimation}
\label{sec:monoculardepth}
Following prior work \cite{zhang2024monst3r,wang2025continuous,wang2025pi}, we use all frames from Sintel, KITTI, and NYUv2 for evaluation. In monocular depth estimation, all models differ only in their scale assumptions.

As shown in \cref{tab:monocular}, \textsc{UniT} performs best in the metric-scale setting and remains competitive under scale-invariant evaluation. We also observe that the offline–online gap narrows in the monocular setting. This is possibly because offline methods rely on multi-frame training with at least two frames, whereas online methods and our \textsc{UniT} can be trained directly from monocular inputs with a single frame.

\begin{table*}[t]
\centering
\ijcvtablelayout{7}

\caption{\rev{Long-horizon perception on the KITTI-Odometry dataset.}}
\label{tab:kitti_odometry}

\begin{tabular}{l ccccc ccccc c c}
\toprule
\multirow{3}{*}{\textbf{Methods}} & \textbf{00} & \textbf{01} & \textbf{02} & \textbf{03} & \textbf{04} & \textbf{05} & \textbf{06} & \textbf{07} & \textbf{08} & \textbf{09} & \textbf{10} & \multirow{3}{*}{\textbf{Avg $\downarrow$}} \\
\cmidrule(lr){2-12}
&  \textit{4542$\times$} &  \textit{1101$\times$} &  \textit{4661$\times$} &  \textit{801$\times$} &  \textit{271$\times$} &  \textit{2761$\times$} &  \textit{1101$\times$} &  \textit{1101$\times$} &  \textit{4071$\times$} &  \textit{1591$\times$} &  \textit{1201$\times$} \\
&  \textit{3.7km} &  \textit{2.5km} &  \textit{5.1km} &  \textit{0.6km} &  \textit{0.4km} &  \textit{2.2km} &  \textit{1.2km} &  \textit{0.7km} &  \textit{3.2km} &  \textit{1.7km} &  \textit{0.9km} \\
\midrule
\revrow StreamVGGT\cite{zhuo2025streaming}        & -    & -    & -    & 146.57 & 82.49 & -    & -    & -   & -    & -    & -    & -  \\
\revrow TTT3R\cite{chen2025ttt3r}        & 173.17 & 542.88 & 283.47 & 108.91 & 13.83 & 137.35 & 103.14 & 69.24 & \textbf{240.28} & 187.35 & \textbf{124.47} & 180.37   \\
\rowcolor{gray!15}
\revrow \textbf{Ours} & \textbf{160.21} & \textbf{119.98} & \textbf{260.20} & \textbf{7.25} & \textbf{1.91} & \textbf{137.27} & \textbf{33.97} & \textbf{39.80} & 248.60 & \textbf{160.81} & 204.62 & \textbf{124.96} \\
\midrule
\revrow VGGT-Long\cite{deng2025vggt}     & 114.01 & 57.16  & 141.83 & 9.00   & 4.52  & 33.65  & 31.82  & 13.71 & 74.32  & 69.41  & 27.83  & 52.48   \\
\rowcolor{gray!15}
\revrow \textbf{Ours-Long}  & \textbf{39.03}  & \textbf{32.92}  & \textbf{115.70} & \textbf{7.00}   & \textbf{2.39}  & \textbf{21.44}  & \textbf{8.92}   & \textbf{8.35}  & \textbf{52.70}  & \textbf{27.90}  & \textbf{9.58}   & \textbf{29.63}   \\
\bottomrule
\end{tabular}
\par\smallskip
\begin{minipage}{\linewidth}\raggedright
\textit{Reported using scale-aligned ATE (m) for comparability with scale-invariant methods. - denotes out of memory (OOM).}
\end{minipage}

\end{table*}

\subsection{Long-Horizon Perception}
\label{sec:longhorizon}
We first evaluate long-horizon perception on the indoor NRGBD dataset with 500 frames. Results are reported for sequence lengths from 50 to 500 frames. To simplify evaluation, we only report metric-scale results here.

\begin{figure}[t]
    \centering
    \begin{minipage}{\linewidth}
        \centering
        \includegraphics[width=\linewidth]{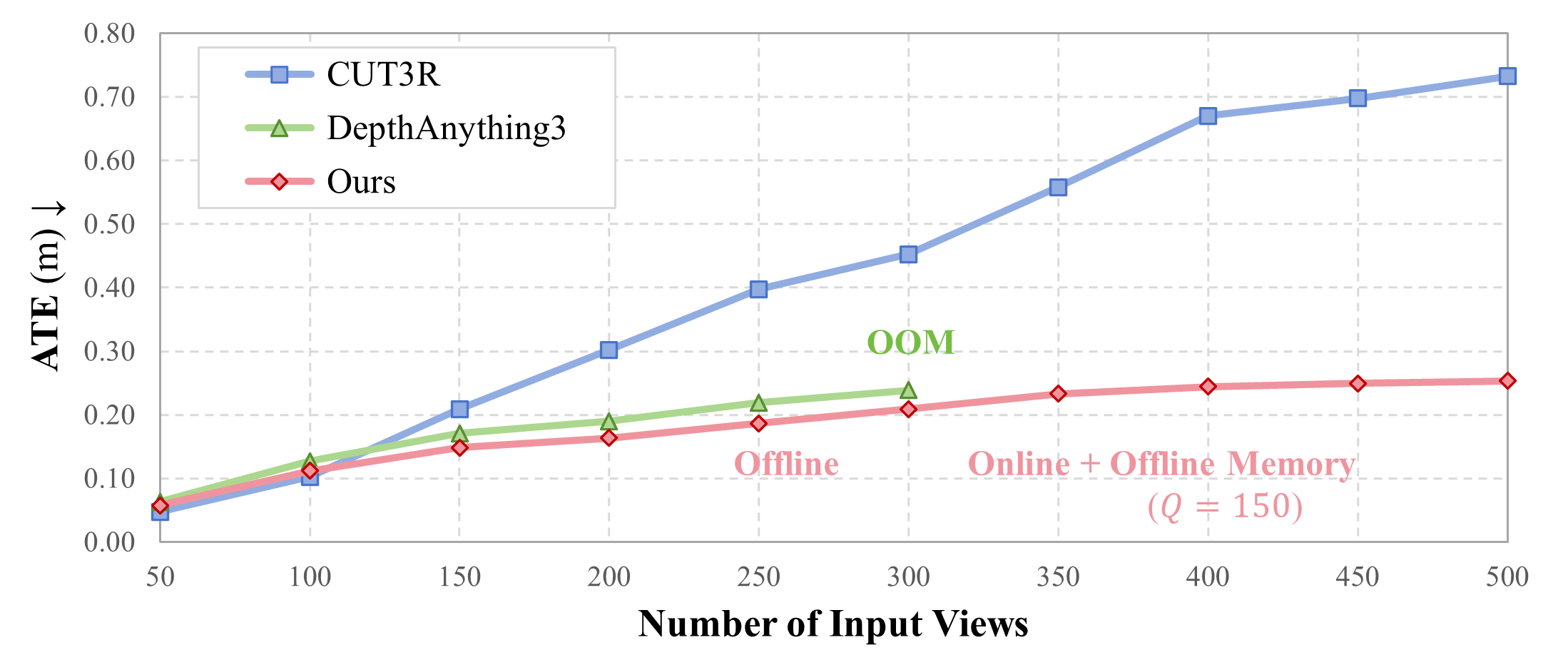}
    \end{minipage}
    \vspace{0.6em}
    \begin{minipage}{\linewidth}
        \centering
        \includegraphics[width=\linewidth]{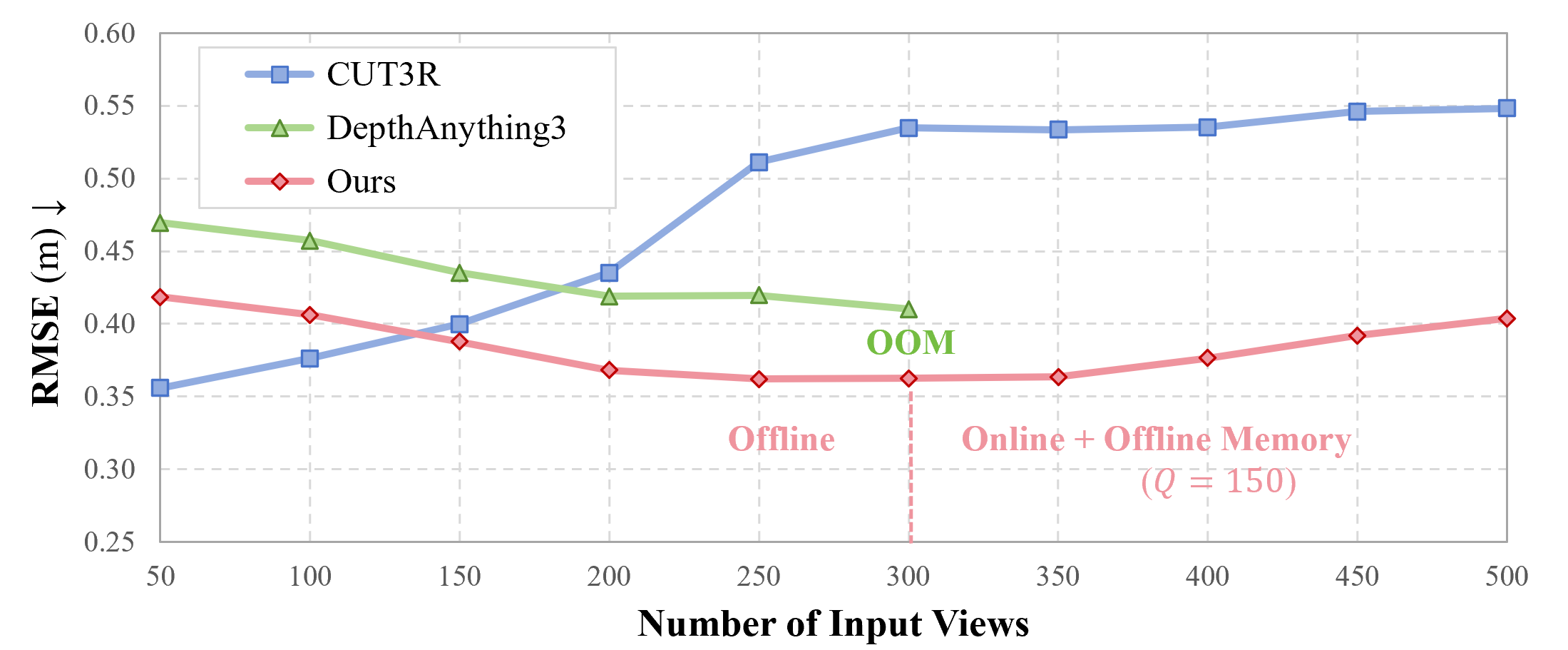}
    \end{minipage}
    \caption{Long-horizon perception on the NRGBD dataset (500 frames). The top plot shows pose accuracy (ATE), and the bottom plot shows depth accuracy (RMSE). All results are evaluated in metric scale.}
    \label{fig:long_horizon}
\end{figure}

In \cref{fig:long_horizon}, we compare \textsc{UniT} with an offline method, DepthAnything3, and an online method, CUT3R. The top plot shows camera pose estimation, and the bottom plot shows video depth estimation. The results indicate that offline methods have a clear advantage before encountering out-of-memory (OOM) issues. For example, in pose estimation with 300 frames, DepthAnything3 achieves approximately half the ATE error of CUT3R, but cannot handle longer sequences due to the quadratic complexity.

Benefiting from the unified formulation, our model naturally supports hybrid offline-online inference. For sequences shorter than 300 frames, we use offline inference, whereas for longer sequences, we switch to online inference. A further advantage is that the online stage can reuse the KV-cache built during the offline stage. Specifically, we first perform offline mode over the initial 150 frames, and then continue with online mode based on the cached memory.

\rev{We further evaluate \textsc{UniT} on the outdoor KITTI-Odometry\cite{geiger2012we} dataset, which contains kilometer-scale sequences. As shown in the top part of \cref{tab:kitti_odometry}, the original \textsc{UniT} outperforms the baselines in almost all scenarios. Meanwhile, we observe performance degradation when generalizing to extremely long sequences exceeding 1,200 frames. We attribute this degradation largely to the train-test length gap. Our model is trained with a limited 24-frame horizon, while inference may involve sequences more than 50$\times$ longer. Training with longer sequences may further reduce this gap.}

\rev{To address this train-test length gap, we further integrate \textsc{UniT} into the VGGT-Long\cite{deng2025vggt} framework, which divides a long trajectory into multiple overlapping short subsequences. Under the same setting as VGGT-Long, our variant (i.e., Ours-Long) further scales to sequences with more than 4,500 frames and 3.7 km, demonstrating the flexibility of \textsc{UniT} for long-horizon perception.}

\begin{table*}[t]
\centering
\ijcvtablelayout{8}

\caption{Metric-scale Multi-view reconstruction results on 7-Scenes, ETH3D, and ScanNetV2 under different modality combinations.}
\label{tab:point_mm}

\begin{tabular}{c c c l c c c c c c c c c c}
\toprule
\multicolumn{3}{c}{\textbf{Modality}} 
& \multirow{2}{*}{\textbf{Methods}} 
& \multicolumn{3}{c}{\textbf{7-Scenes}} 
& \multicolumn{3}{c}{\textbf{ETH3D}} 
& \multicolumn{3}{c}{\textbf{ScanNetV2}} 
& \multirow{2}{*}{\textbf{Rank $\downarrow$}} \\
\cmidrule(lr){1-3}
\cmidrule(lr){5-7}
\cmidrule(lr){8-10}
\cmidrule(lr){11-13}
$\mathbf{K}$ & $\mathbf{[R|T]}$ & $\mathbf{D}$
& 
& Acc.$\downarrow$ & Comp.$\downarrow$ & N.C.$\uparrow$
& Acc.$\downarrow$ & Comp.$\downarrow$ & N.C.$\uparrow$
& Acc.$\downarrow$ & Comp.$\downarrow$ & N.C.$\uparrow$
& \\
\midrule

\multirow{3}{*}{$\checkmark$} & \multirow{3}{*}{} & \multirow{3}{*}{}
& MapAnything\cite{keetha2025mapanything} & 0.399 & 0.159 & 0.589 & 0.651 & \textbf{0.454} & 0.753 & 0.116 & 0.072 & 0.709 & 2.78 \\
& & 
& DepthAnything3\cite{lin2025depthany3}   & \underline{0.074} & \underline{0.089} & \underline{0.719} & \textbf{0.396} & \underline{0.464} & \textbf{0.809} & \underline{0.032} & \underline{0.042} & \underline{0.780} & \underline{1.78} \\
& & 
& \cellcolor{gray!15}\textbf{Ours($G\!=\!N$)} & \cellcolor{gray!15}\textbf{0.042} & \cellcolor{gray!15}\textbf{0.041} & \cellcolor{gray!15}\textbf{0.752} & \cellcolor{gray!15}\underline{0.486} & \cellcolor{gray!15}0.542 & \cellcolor{gray!15}\underline{0.801} & \cellcolor{gray!15}\textbf{0.022} & \cellcolor{gray!15}\textbf{0.026} & \cellcolor{gray!15}\textbf{0.804} & \cellcolor{gray!15}\textbf{1.44} \\
\midrule

\multirow{3}{*}{} & \multirow{3}{*}{$\checkmark$} & \multirow{3}{*}{}
& MapAnything\cite{keetha2025mapanything} & \underline{0.052} & \underline{0.042} & 0.727 & \underline{0.260} & \underline{0.201} & 0.841 & 0.094 & 0.097 & 0.701 & 2.56 \\
& & 
& DepthAnything3\cite{lin2025depthany3}   & 0.144 & 0.042 & \underline{0.774} & 0.390 & 0.236 & \underline{0.864} & \underline{0.036} & \underline{0.043} & \underline{0.781} & \underline{2.44} \\
& & 
& \cellcolor{gray!15}\textbf{Ours($G\!=\!N$)} & \cellcolor{gray!15}\textbf{0.028} & \cellcolor{gray!15}\textbf{0.028} & \cellcolor{gray!15}\textbf{0.796} & \cellcolor{gray!15}\textbf{0.140} & \cellcolor{gray!15}\textbf{0.115} & \cellcolor{gray!15}\textbf{0.878} & \cellcolor{gray!15}\textbf{0.022} & \cellcolor{gray!15}\textbf{0.025} & \cellcolor{gray!15}\textbf{0.803} & \cellcolor{gray!15}\textbf{1.00} \\
\midrule

\multirow{3}{*}{} & \multirow{3}{*}{} & \multirow{3}{*}{$\checkmark$}
& MapAnything\cite{keetha2025mapanything} & 0.078 & \underline{0.076} & 0.725 & 0.478 & 0.388 & 0.788 & 0.078 & 0.078 & 0.740 & 2.89 \\
& & 
& DepthAnything3\cite{lin2025depthany3}   & \underline{0.068} & 0.080 & \underline{0.741} & \underline{0.390} & \textbf{0.237} & \textbf{0.864} & \underline{0.028} & \underline{0.034} & \underline{0.791} & \underline{1.89} \\
& & 
& \cellcolor{gray!15}\textbf{Ours($G\!=\!N$)} & \cellcolor{gray!15}\textbf{0.031} & \cellcolor{gray!15}\textbf{0.034} & \cellcolor{gray!15}\textbf{0.774} & \cellcolor{gray!15}\textbf{0.316} & \cellcolor{gray!15}\underline{0.283} & \cellcolor{gray!15}\underline{0.842} & \cellcolor{gray!15}\textbf{0.016} & \cellcolor{gray!15}\textbf{0.020} & \cellcolor{gray!15}\textbf{0.813} & \cellcolor{gray!15}\textbf{1.22} \\
\midrule

\multirow{3}{*}{$\checkmark$} & \multirow{3}{*}{$\checkmark$} & \multirow{3}{*}{}
& MapAnything\cite{keetha2025mapanything} & 0.043 & 0.038 & 0.741 & 0.236 & 0.192 & 0.852 & 0.059 & 0.059 & 0.729 & 3.00 \\
& & 
& DepthAnything3\cite{lin2025depthany3}   & \textbf{0.021} &  \underline{0.026} & \textbf{0.804} & \textbf{0.093} & \textbf{0.065} & \textbf{0.918} &  \underline{0.032} &  \underline{0.041} &  \underline{0.787} & \textbf{1.33} \\
& & 
& \cellcolor{gray!15}\textbf{Ours($G\!=\!N$)} & \cellcolor{gray!15} \underline{0.024} & \cellcolor{gray!15}\textbf{0.026} & \cellcolor{gray!15} \underline{0.802} & \cellcolor{gray!15} \underline{0.147} & \cellcolor{gray!15} \underline{0.120} & \cellcolor{gray!15} \underline{0.884} & \cellcolor{gray!15}\textbf{0.020} & \cellcolor{gray!15}\textbf{0.024} & \cellcolor{gray!15}\textbf{0.807} & \cellcolor{gray!15}\underline{1.67} \\
\midrule

\multirow{3}{*}{$\checkmark$} & \multirow{3}{*}{} & \multirow{3}{*}{$\checkmark$}
& MapAnything\cite{keetha2025mapanything} & \underline{0.046} & \underline{0.059} & \underline{0.755} & 0.480 & \underline{0.339} & 0.798 & \underline{0.024} & \underline{0.025} & \underline{0.793} & \underline{2.22} \\
& & 
& DepthAnything3\cite{lin2025depthany3}   & 0.069 & 0.079 & 0.744 & \underline{0.403} & 0.390 & \underline{0.831} & 0.028 & 0.034 & 0.792 & 2.78 \\
& & 
& \cellcolor{gray!15}\textbf{Ours($G\!=\!N$)} & \cellcolor{gray!15}\textbf{0.028} & \cellcolor{gray!15}\textbf{0.032} & \cellcolor{gray!15}\textbf{0.782} & \cellcolor{gray!15}\textbf{0.317} & \cellcolor{gray!15}\textbf{0.286} & \cellcolor{gray!15}\textbf{0.841} & \cellcolor{gray!15}\textbf{0.015} & \cellcolor{gray!15}\textbf{0.020} & \cellcolor{gray!15}\textbf{0.815} & \cellcolor{gray!15}\textbf{1.00} \\
\midrule

\multirow{3}{*}{} & \multirow{3}{*}{$\checkmark$} & \multirow{3}{*}{$\checkmark$}
& MapAnything\cite{keetha2025mapanything} & \textbf{0.017} & \textbf{0.016} & \underline{0.798} & 0.126 & \underline{0.091} & 0.867 & 0.039 & \underline{0.024} & 0.779 & 2.22 \\
& & 
& DepthAnything3\cite{lin2025depthany3}   & 0.058 & 0.080 & 0.723 & \underline{0.108} & \textbf{0.079} & \textbf{0.903} & \underline{0.027} & 0.031 & \underline{0.799} & \underline{2.11} \\
& & 
& \cellcolor{gray!15}\textbf{Ours($G\!=\!N$)} & \cellcolor{gray!15}\underline{0.021} & \cellcolor{gray!15}\underline{0.023} & \cellcolor{gray!15}\textbf{0.812} & \cellcolor{gray!15}\textbf{0.108} & \cellcolor{gray!15}0.093 & \cellcolor{gray!15}\underline{0.888} & \cellcolor{gray!15}\textbf{0.014} & \cellcolor{gray!15}\textbf{0.018} & \cellcolor{gray!15}\textbf{0.817} & \cellcolor{gray!15}\textbf{1.67} \\
\midrule

\multirow{3}{*}{$\checkmark$} & \multirow{3}{*}{$\checkmark$} & \multirow{3}{*}{$\checkmark$}
& MapAnything\cite{keetha2025mapanything} & \textbf{0.011} & \textbf{0.012} & \underline{0.818} & \underline{0.104} & \underline{0.074} & 0.880 & \textbf{0.009} & \textbf{0.010} & \textbf{0.823} & \textbf{1.44} \\
& & 
& DepthAnything3\cite{lin2025depthany3}   & 0.021 & 0.025 & 0.804 & \textbf{0.089} & \textbf{0.061} & \textbf{0.925} & 0.027 & 0.033 & 0.797 & 2.33 \\
& & 
& \cellcolor{gray!15}\textbf{Ours($G\!=\!N$)} & \cellcolor{gray!15}\underline{0.018} & \cellcolor{gray!15}\underline{0.022} & \cellcolor{gray!15}\textbf{0.818} & \cellcolor{gray!15}0.105 & \cellcolor{gray!15}0.087 & \cellcolor{gray!15}\underline{0.897} & \cellcolor{gray!15}\underline{0.014} & \cellcolor{gray!15}\underline{0.018} & \cellcolor{gray!15}\underline{0.820} & \cellcolor{gray!15}\underline{2.22} \\
\bottomrule
\end{tabular}

\end{table*}

\subsection{Multi-Modal Reconstruction}
\label{sec:multimodal}
In this subsection, frames are sampled with strides of 200, 5, and 20 on 7-Scenes, ETH3D, and ScanNetv2, respectively. The optional modalities include depth maps $\mathbf{D}$, camera intrinsics $\mathbf{K}$, and camera extrinsics $\mathbf{[R|T]}$. We only report metric-scale results here for simplicity.

As shown in \cref{tab:point_mm}, \textsc{UniT} attains the best performance in most multi-modal combinations, highlighting its strong flexibility in supporting auxiliary modalities. We note that MapAnything performs better when all modalities are available. A possible explanation is that it is trained from scratch with multi-modal inputs, which may make it better suited to fully exploit complete modal observations.

\begin{table*}[t]
\centering
\ijcvtablelayout{9}
\caption{Metric-scale depth completion on Sintel, KITTI, and NYUv2 under different sparse patterns.}
\label{tab:depthcompletion}

\begin{tabular}{c l c c c c c c c c c c}
\toprule
\multirow{2}{*}{\textbf{Pattern}} 
& \multirow{2}{*}{\textbf{Methods}}
& \multicolumn{3}{c}{\textbf{Sintel}} 
& \multicolumn{3}{c}{\textbf{KITTI}} 
& \multicolumn{3}{c}{\textbf{NYUv2}} 
& \multirow{2}{*}{\textbf{Rank$\downarrow$}} \\

\cmidrule(lr){3-5}
\cmidrule(lr){6-8}
\cmidrule(lr){9-11}

& 
& AbsRel$\downarrow$ & RMSE$\downarrow$ & $\delta\!<\!1.25$$\uparrow$
& AbsRel$\downarrow$ & RMSE$\downarrow$ & $\delta\!<\!1.25$$\uparrow$
& AbsRel$\downarrow$ & RMSE$\downarrow$ & $\delta\!<\!1.25$$\uparrow$
& \\
\midrule

\multirow{3}{*}{\textit{Uniform}}
& MapAnything\cite{keetha2025mapanything} & \underline{0.231} & \textbf{3.613} & \textbf{0.819} & 0.078 & 3.563 & 0.947 & \underline{0.093} & \underline{0.283} & \underline{0.953} & \underline{2.11} \\
& DepthAnything3\cite{lin2025depthany3}   & 0.310 & 5.146 & 0.614 & \underline{0.075} & \textbf{3.215} & \underline{0.957} & 0.098 & 0.546 & 0.893 & 2.56 \\
& \cellcolor{gray!15}\textbf{Ours}   & \cellcolor{gray!15}\textbf{0.217} & \cellcolor{gray!15}\underline{4.680} & \cellcolor{gray!15}\underline{0.710} & \cellcolor{gray!15}\textbf{0.059} & \cellcolor{gray!15}\underline{3.546} & \cellcolor{gray!15}\textbf{0.966} & \cellcolor{gray!15}\textbf{0.046} & \cellcolor{gray!15}\textbf{0.269} & \cellcolor{gray!15}\textbf{0.963} & \cellcolor{gray!15}\textbf{1.33} \\
\midrule

\multirow{3}{*}{\textit{LiDAR}}
& MapAnything\cite{keetha2025mapanything} & \underline{0.239} & \textbf{3.933} & \textbf{0.802} & 0.089 & 3.991 & 0.934 & \underline{0.077} & \underline{0.299} & \underline{0.962} & \underline{2.00} \\
& DepthAnything3\cite{lin2025depthany3}   & 0.325 & 5.220 & 0.613 & \underline{0.079} & \textbf{3.232} & \underline{0.951} & 0.092 & 0.542 & 0.899 & 2.56 \\
& \cellcolor{gray!15}\textbf{Ours}   & \cellcolor{gray!15}\textbf{0.220} & \cellcolor{gray!15}\underline{4.786} & \cellcolor{gray!15}\underline{0.706} & \cellcolor{gray!15}\textbf{0.059} & \cellcolor{gray!15}\underline{3.555} & \cellcolor{gray!15}\textbf{0.965} & \cellcolor{gray!15}\textbf{0.047} & \cellcolor{gray!15}\textbf{0.274} & \cellcolor{gray!15}\textbf{0.962} & \cellcolor{gray!15}\textbf{1.44} \\
\midrule

\multirow{3}{*}{\textit{SfM}}
& MapAnything\cite{keetha2025mapanything} & 0.380 & 5.713 & 0.547 & 0.102 & 4.242 & 0.925 & 0.101 & \underline{0.356} & \underline{0.902} & 2.78 \\
& DepthAnything3\cite{lin2025depthany3}   & \underline{0.350} & \underline{5.483} & \underline{0.564} & \underline{0.087} & \textbf{3.328} & \underline{0.942} & \underline{0.096} & 0.546 & 0.896 & \underline{2.11} \\
& \cellcolor{gray!15}\textbf{Ours}   & \cellcolor{gray!15}\textbf{0.265} & \cellcolor{gray!15}\textbf{5.192} & \cellcolor{gray!15}\textbf{0.633} & \cellcolor{gray!15}\textbf{0.061} & \cellcolor{gray!15}\underline{3.583} & \cellcolor{gray!15}\textbf{0.965} & \cellcolor{gray!15}\textbf{0.050} & \cellcolor{gray!15}\textbf{0.277} & \cellcolor{gray!15}\textbf{0.960} & \cellcolor{gray!15}\textbf{1.11} \\
\midrule

\multirow{3}{*}{\textit{Grid}}
& MapAnything\cite{keetha2025mapanything} & \underline{0.253} & \textbf{3.796} & \textbf{0.780} & 0.095 & 4.188 & 0.925 & \underline{0.087} & \underline{0.308} & \underline{0.932} & \underline{2.11} \\
& DepthAnything3\cite{lin2025depthany3}   & 0.317 & 5.172 & 0.611 & \underline{0.081} & \textbf{3.286} & \underline{0.948} & 0.099 & 0.548 & 0.892 & 2.56 \\
& \cellcolor{gray!15}\textbf{Ours}   & \cellcolor{gray!15}\textbf{0.230} & \cellcolor{gray!15}\underline{4.742} & \cellcolor{gray!15}\underline{0.701} & \cellcolor{gray!15}\textbf{0.060} & \cellcolor{gray!15}\underline{3.589} & \cellcolor{gray!15}\textbf{0.965} & \cellcolor{gray!15}\textbf{0.047} & \cellcolor{gray!15}\textbf{0.269} & \cellcolor{gray!15}\textbf{0.963} & \cellcolor{gray!15}\textbf{1.33} \\
\bottomrule
\end{tabular}

\end{table*}

\begin{table}[t]
\centering
\ijcvtablelayout{10}
\caption{Ablation study on modal attention components.}
\label{tab:ablation_modalattn}

\begin{tabular}{c c c c c c c c}
\toprule
\multicolumn{3}{c}{\textbf{Modal Attention}} & \multicolumn{2}{c}{Image-Only} & \multicolumn{2}{c}{Multi-Modal} & \multirow{2}{*}{\textbf{Avg$\downarrow$}}\\
\cmidrule(lr){1-3} \cmidrule(lr){4-5} \cmidrule(lr){6-7}

CrossAttn & Concat. & 4~stages & \textit{S-Inv.}& \textit{M-Sca.} & \textit{S-Inv.}& \textit{M-Sca.} &\\
\midrule
$\times$      & $\times$      & $\checkmark$      & 0.050   & 0.117   & 0.037   & 0.092 & 0.074\\
$\checkmark$  & $\times$      & $\checkmark$      & \textbf{0.040}   & 0.109   & 0.036   & 0.095 & 0.070\\
$\checkmark$  & $\checkmark$  & $\times$          & 0.048   & \textbf{0.107}   & 0.035   & 0.088 & 0.069\\
\rowcolor{gray!15}
$\checkmark$  & $\checkmark$  & $\checkmark$  & 0.046 & 0.108 & \textbf{0.033}   & \textbf{0.088} & \textbf{0.068}\\
\bottomrule
\end{tabular}
\par\smallskip
\begin{minipage}{\linewidth}\raggedright
\textit{Reported using $(\text{Acc.}\! +\! \text{Comp.})/2$ on 7-Scenes, NRGBD, and DTU.}
\end{minipage}

\end{table}

\subsection{Depth Completion}
\label{sec:depthcompletion}
This subsection follows the same data setting as monocular depth estimation, but additionally provides sparse depth maps with four sampling patterns as sensor prompts. As in prior work \cite{wang2025pacgdc,zuo2025omni,long2024sparsedc}, the four patterns include uniform sampling with random densities from 0\% to 100\%, random LiDAR patterns from 1 to 128 beams, SfM feature points extracted using SIFT descriptors, and super-resolution grid patterns with random downsampling factors from 1 to 16. We only report metric-scale results here for simplicity.

In \cref{tab:depthcompletion}, \textsc{UniT} ranks first across all evaluated scenarios and remains robust across different sparsity patterns. This advantage stems in part from our goal of unified geometry learning, where such variations must be explicitly taken into account. Accordingly, we simulate multiple sparse patterns with different densities during training, which helps reduce the train-test gap across diverse sensor configurations.

\begin{table}[t]
\centering
\ijcvtablelayout{11}

\caption{Ablation study on loss configurations.}
\label{tab:ablation_loss}

\begin{tabular}{l c c c c c}
\toprule
\multirow{2}{*}{\textbf{Loss Configuration}} & \multicolumn{2}{c}{Offline} & \multicolumn{2}{c}{Online} & \multirow{2}{*}{\textbf{Avg$\downarrow$}}\\
\cmidrule(lr){2-3} \cmidrule(lr){4-5}
& \textit{S-Inv.} & \textit{M-Sca.} & \textit{S-Inv.}& \textit{M-Sca.} & \\
\midrule
\textbf{a):} Direct Regression & 0.053 & 0.201 & \textbf{0.067} & 0.433 & 0.188\\
\textbf{b):} Scale-Adaptive Design & \textbf{0.044} & 0.108 & 0.069 & 0.147 & 0.092 \\
\rowcolor{gray!15}
\textbf{c):} b) + Shuffled Normal & 0.046 & \textbf{0.108} & 0.070 & \textbf{0.141} & \textbf{0.091} \\
\bottomrule
\end{tabular}
\par\smallskip
\begin{minipage}{\linewidth}\raggedright
\textit{Reported using $(\text{Acc.}\! +\! \text{Comp.})/2$ on 7-Scenes, NRGBD, and DTU. }
\end{minipage}

\end{table}

\begin{table}[t]
\centering
\ijcvtablelayout{12}
\caption{Ablation study of KV-cache drop strategies.}
\label{tab:ablation_cache}

\begin{tabular}{l c c c c c c}
\toprule
\multirow{2}{*}{\textbf{Strategy}} &
\multicolumn{3}{c}{\textit{Scale-Invariant}} &
\multicolumn{3}{c}{\textit{Metric-Scale}} \\
\cmidrule(lr){2-4} \cmidrule(lr){5-7}
& ATE$\downarrow$ & RPE\textsubscript{tra}$\downarrow$ & RPE\textsubscript{rot}$\downarrow$ & ATE$\downarrow$ & RPE\textsubscript{tra}$\downarrow$ & RPE\textsubscript{rot}$\downarrow$\\
\midrule
Full Cache                 & 0.037 & 0.016 & 0.424 & 0.045 & 0.016 & 0.424 \\
\arrayrulecolor{gray}
\midrule
First-in-First-out                     & 0.047 & 0.018 & 0.470 & 0.057 & 0.018 & 0.470 \\
Random Drop                  & 0.041 & 0.017 & 0.434 & 0.049 & 0.017 & 0.434 \\
Token Merge                   & 0.041 & 0.017 & 0.437 & 0.050 & 0.017 & 0.437\\
\rowcolor{gray!15}
\textbf{Stride Drop}                  & \textbf{0.038} & \textbf{0.017} & \textbf{0.427} & \textbf{0.046} & \textbf{0.017} & \textbf{0.427}\\
\bottomrule
\end{tabular}
\par\smallskip
\begin{minipage}{\linewidth}\raggedright
\textit{Reported on ScanNetV2 with 90 frames using queue capacity $Q=50$.}
\end{minipage}

\end{table}

\subsection{Ablation Study}
\label{sec:ablation}

\rev{In this section, we first ablate the design choices to determine our final model under the same data setting.} To reduce the computational cost, these models are trained at a resolution of 244 for 60K iterations. We then analyze the characteristics of our final model in the remaining experiments.

\noindent \textbf{Modal Attention.} We ablate the modal attention in \cref{tab:ablation_modalattn} under both image-only and multi-modal settings. The first row replaces modal attention with simple linear projections at four stages, resulting in substantial performance degradation. The second row shows that the concatenation operation within modal attention is beneficial, as it explicitly establishes spatial correspondence between modalities. The third row further indicates that injecting multi-modal prompts at multiple stages consistently improves overall performance.

\noindent \textbf{Loss Function.} We ablate the loss components in \cref{tab:ablation_loss} under both offline and online settings. As shown in the first row, directly applying the $\ell_1$ regression loss in metric-scale space, as in \citet{wang2025continuous}, results in a clear drop in metric-scale performance. To mitigate this issue, we introduce the scale-adaptive design described in \cref{sec:metricscale}, which substantially improves convergence, as evidenced by the second row. Additionally, the third row demonstrates that the shuffled normal loss serves as an effective global geometric regularizer.

\noindent \textbf{KV-Cache Drop Strategy.} We compare four simple strategies for removing outdated tokens when the queue capacity is exceeded in our queue-style KV caching: first-in-first-out, random dropping, token merging via neighbor interpolation, and stride-based dropping. As shown in \cref{tab:ablation_cache}, stride-based dropping consistently yields the best performance. More importantly, these results confirm that our model can safely discard outdated cached tokens during inference, enabling bounded memory usage over time.

\begin{figure}[!t]
    \centering
    \begin{minipage}{\linewidth}
        \centering
        \includegraphics[width=\linewidth]{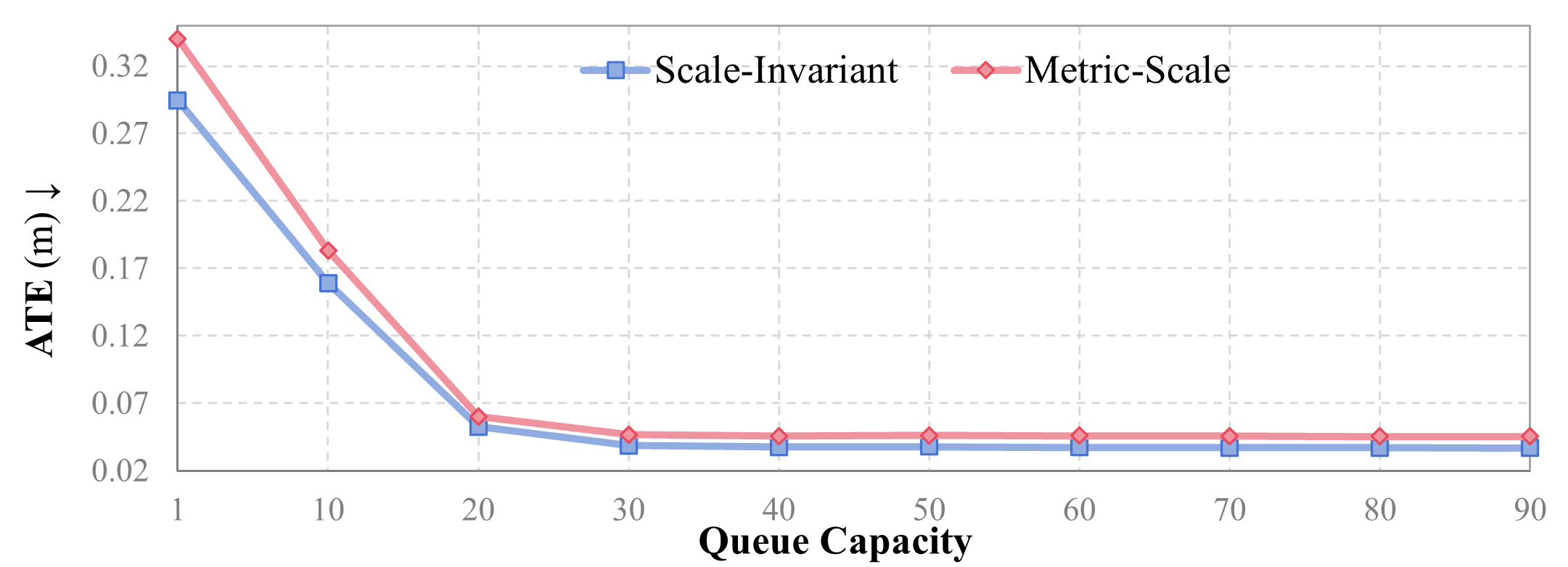}
    \end{minipage}
    \vspace{0.6em}
    \begin{minipage}{\linewidth}
        \centering
        \includegraphics[width=\linewidth]{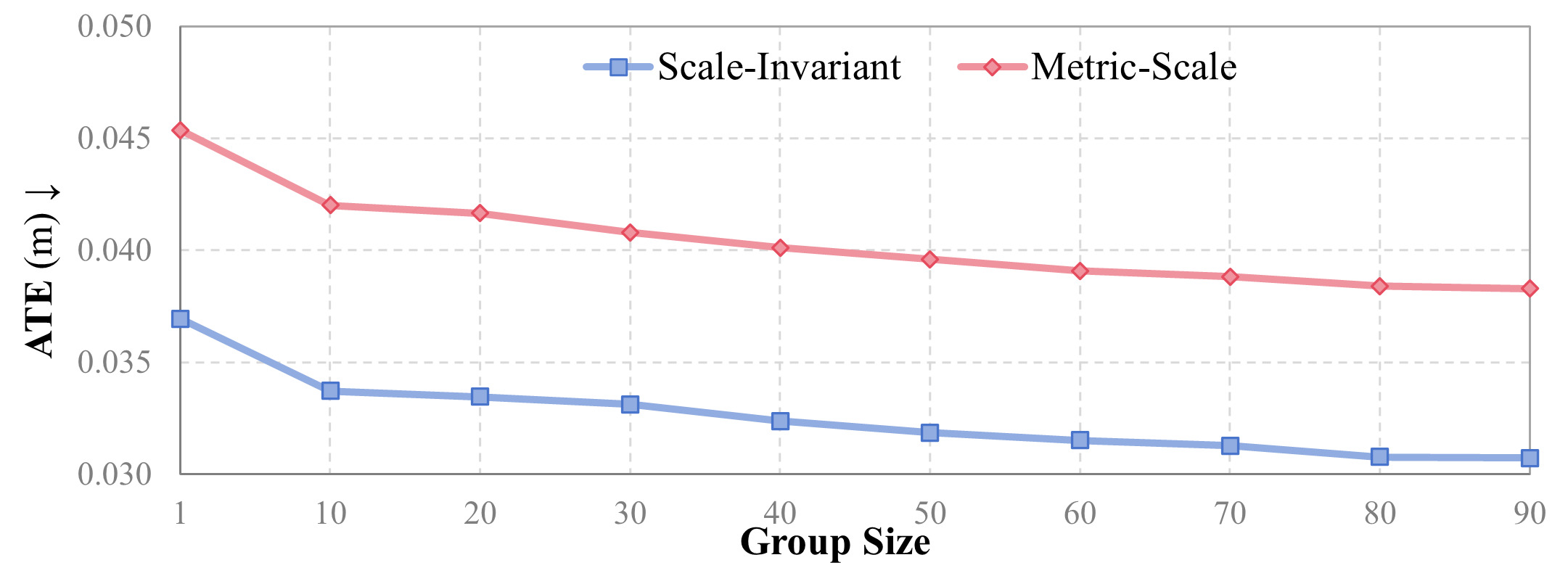}
    \end{minipage}
    \caption{Ablation study of KV-cache queue capacity and group size for camera pose estimation on ScanNetV2 \rev{(100 scenes, 90 frames each)}.}
    \label{fig:ablation_inference}
\end{figure}

\begin{figure}[!t]
    \centering
    \includegraphics[width=\linewidth]{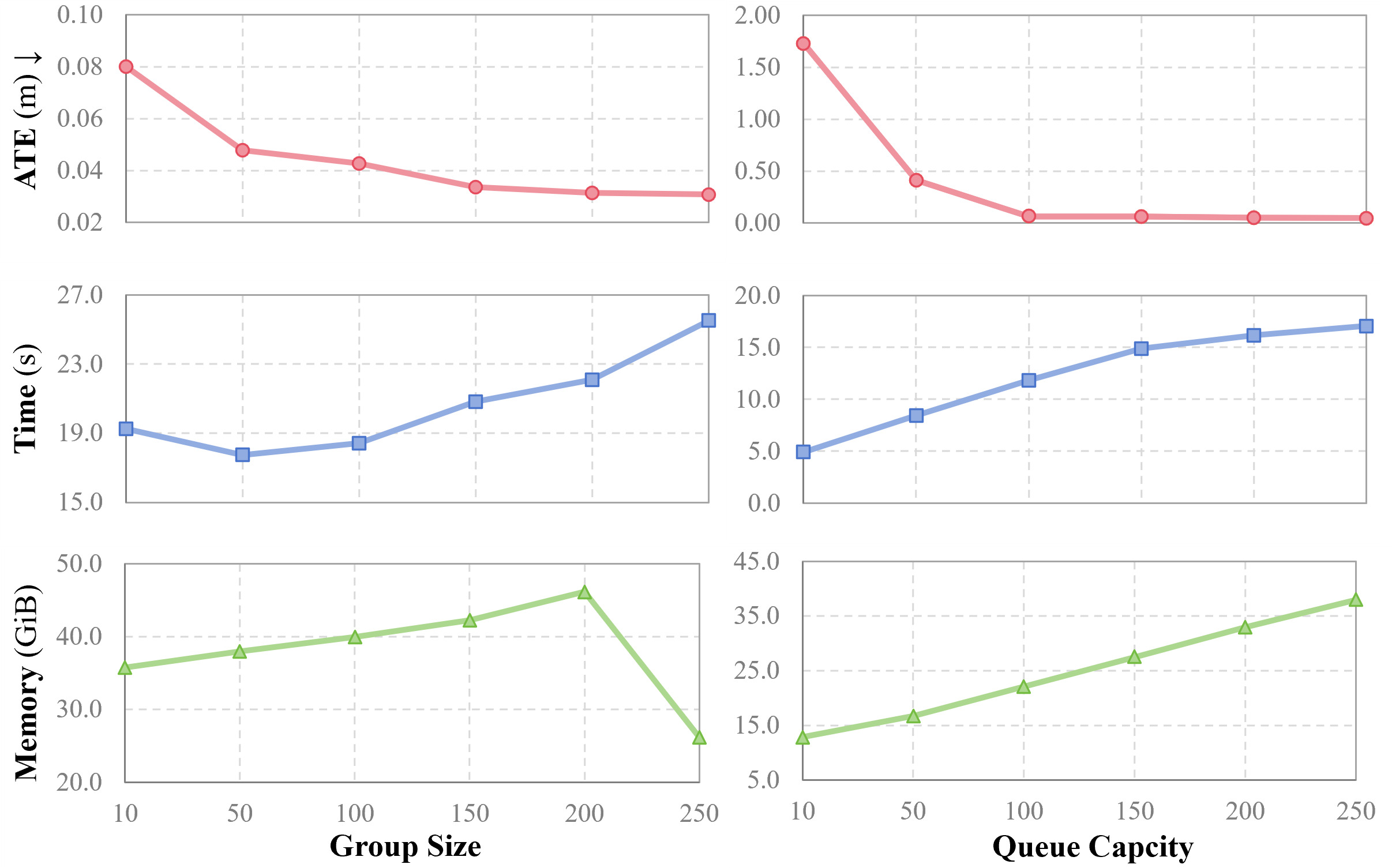}
    \caption{\rev{Accuracy, runtime, and memory usage under different group sizes and queue capacities on the NRGBD test set, evaluated on 250 frames sampled with a stride of 4 using a single H100.}}
    \label{fig:cost_variants}
\end{figure}

\noindent \textbf{Effect of Queue Capacity.} In the top plot of \cref{fig:ablation_inference}, we study the effect of different queue capacities, ranging from a minimum of 1 to a maximum of 90. The results show a clear trend that larger queue capacities consistently lead to better performance. Meanwhile, a capacity of $N/3$ achieves a favorable balance between performance and efficiency.

\noindent \textbf{Effect of Group Size.} Our unified formulation enables our model to comprehensively accommodate different view configurations by varying the group size. As shown in the bottom plot of \cref{fig:ablation_inference}, larger group sizes consistently improve performance, since more frames can interact through bidirectional attention within each group. Meanwhile, our model maintains stable results across a broad range of view configurations, highlighting the robustness of our unified design.

\noindent \rev{\textbf{Accuracy vs. Efficiency.}
We further analyze the joint effects of group size and queue capacity on accuracy, runtime, and memory consumption in \cref{fig:cost_variants}. Overall, increasing either the group size or queue capacity generally improves accuracy at the cost of higher computation. The memory drop in \cref{fig:cost_variants} occurs because streaming mode (i.e., group size $< 250$) incurs additional memory overhead for KV-cache storage, whereas offline mode (i.e., group size $=250$) avoids this but requires more runtime. We thus recommend offline mode for short-to-moderate sequences and streaming for long ones.}

\noindent \rev{\textbf{Unified vs. Specialized Learning.} We compare our unified model against three specialized variants, including offline-only, online-only, and image-only. These variants are derived by modifying attention masks and disabling modal attention accordingly. As shown in \cref{tab:ablation_unified}, the unified model achieves the best overall performance, while remaining competitive with each specialized variant on its respective setting.}

\begin{table}[t]
\centering
\ijcvtablelayout{13}

\caption{\rev{Ablation study on three specialized variants.}}
\label{tab:ablation_unified}

\begin{tabular}{l c c c c c c c}
\toprule
\multirow{2}{*}{\textbf{Strategy}} & \multicolumn{2}{c}{Offline-Only} & \multicolumn{2}{c}{Online-Only} & \multicolumn{2}{c}{Image-Only} & \multirow{2}{*}{\textbf{Avg$\downarrow$}}\\
\cmidrule(lr){2-3} \cmidrule(lr){4-5} \cmidrule(lr){6-7}
& \textit{S-Inv.} & \textit{M-Sca.} & \textit{S-Inv.}& \textit{M-Sca.} & \textit{S-Inv.}& \textit{M-Sca.} & \\
\midrule
\revrow
Specialized & 0.052 & 0.113 & \textbf{0.060} & 0.184 & 0.046 & \textbf{0.097} & 0.092 \\
\rowcolor{gray!15}
\revrow
Unified & \textbf{0.046} & \textbf{0.108} & 0.070 & \textbf{0.141} & \textbf{0.046} & 0.108 & \textbf{0.087} \\
\bottomrule
\end{tabular}
\par\smallskip
\begin{minipage}{\linewidth}\raggedright
\textit{Reported using $(\text{Acc.}\! +\! \text{Comp.})/2$ on 7-Scenes, NRGBD, and DTU. }
\end{minipage}

\end{table}

\begin{figure}[t]
    \centering
    \includegraphics[width=\linewidth]{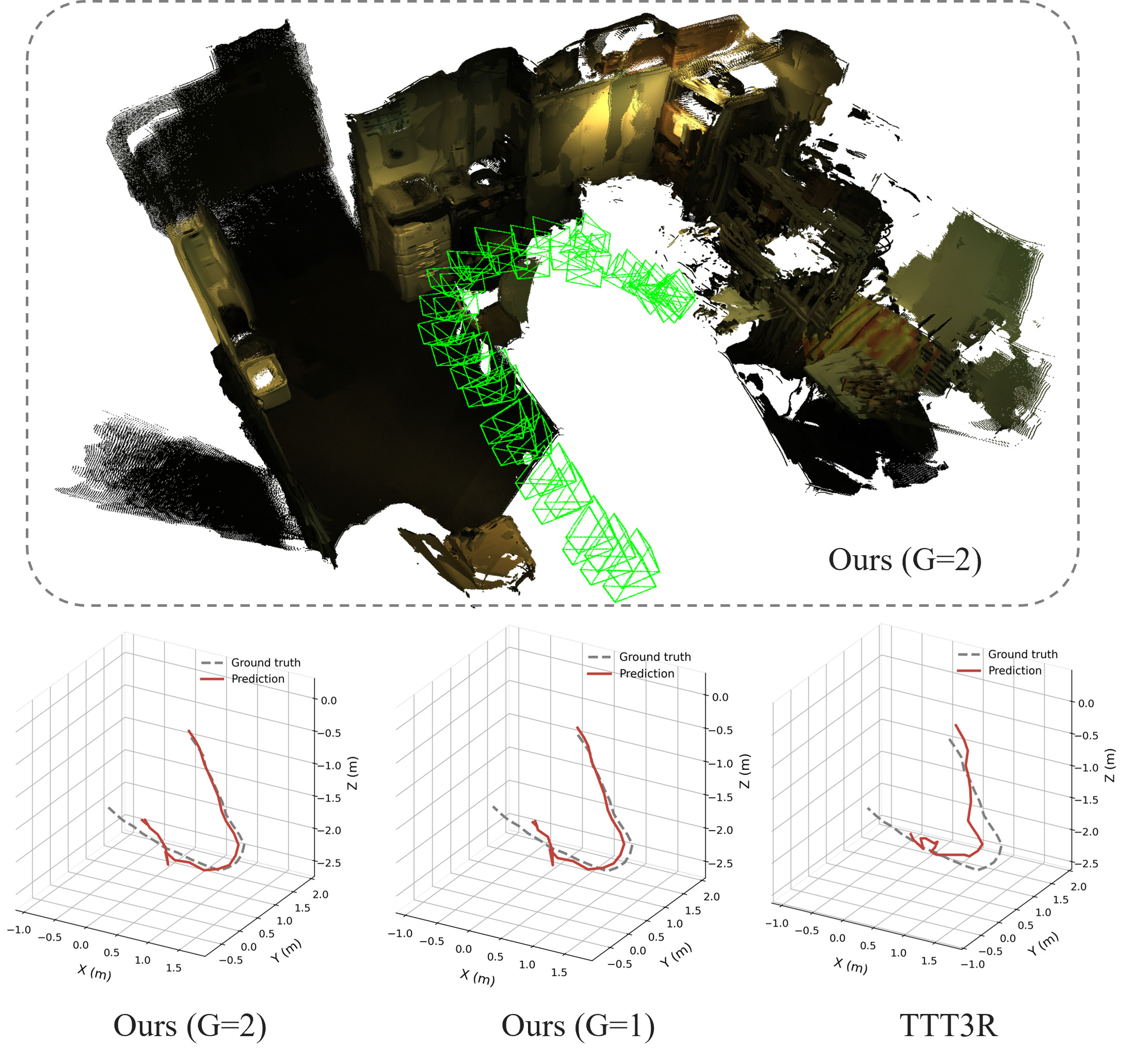}
    \caption{\rev{Qualitative results for binocular streaming from the ETH3D-SLAM~\cite{schops2019bad} dataset. The trajectory is visualized using the midpoint between the two views at each timestep.}}
    \label{fig:binocular}
\end{figure}

\noindent \rev{\textbf{Binocular Streaming.} We further investigate whether increasing the group size can better support multi-view streaming. We evaluate our model with group sizes of 1, corresponding to monocular streaming, and 2, corresponding to binocular streaming, on laboratory sequences from ETH3D-SLAM \cite{schops2019bad}. As shown in \cref{tab:binocular}, binocular streaming outperforms monocular streaming on ATE and RPE, including TTT3R \cite{chen2025ttt3r} and our model with a group size of 1. The qualitative results in \cref{fig:binocular} further demonstrate that our model can effectively operate in binocular streaming, even under challenging low-light conditions.}

\begin{table}[t]
\centering
\ijcvtablelayout{14}
\caption{\rev{Ablation study on binocular streaming scenarios.}}
\label{tab:binocular}

\begin{tabular}{l c c c c c c}
\toprule
\multirow{2}{*}{\textbf{Strategy}} &
\multicolumn{3}{c}{\textit{Scale-Invariant}} &
\multicolumn{3}{c}{\textit{Metric-Scale}} \\
\cmidrule(lr){2-4} \cmidrule(lr){5-7}
& ATE$\downarrow$ & RPE\textsubscript{tra}$\downarrow$ & RPE\textsubscript{rot}$\downarrow$ & ATE$\downarrow$ & RPE\textsubscript{tra}$\downarrow$ & RPE\textsubscript{rot}$\downarrow$\\
\midrule
\revrow TTT3R\cite{chen2025ttt3r}   & 0.548 & 0.316 & \textbf{6.059} & 0.645 & 0.236 & \textbf{6.059} \\
\revrow \textbf{Ours($G=1$)}      & 0.374 & 0.201 & 7.067 & 0.395 & 0.206 & 7.067 \\
\rowcolor{gray!15}
\revrow \textbf{Ours($G=2$)}      & \textbf{0.347} & \textbf{0.197} & 6.994 & \textbf{0.363} & \textbf{0.193} & 6.994 \\
\bottomrule
\end{tabular}

\end{table}

\section{Conclusion}
This paper presents \textsc{UniT}, a feed-forward model for unified geometry learning. Built upon the proposed group causal transformer, it reformulates a broad spectrum of geometric perception capabilities within a simple yet powerful framework. \textsc{UniT} comprehensively accommodates arbitrary view configurations and modality combinations, while supporting metric-scale estimation and long-horizon scalability. Extensive experiments demonstrate that \textsc{UniT} serves as a powerful 3D foundation model, effectively supporting diverse tasks, such as multi-view reconstruction, camera pose estimation, and video and monocular depth estimation, long-horizon perception, multi-modal reconstruction, and depth completion.

\rev{\textsc{UniT} provides a basic framework for unified geometry learning. As future extensions, it would be promising to incorporate cache compression techniques, optimization-based pipelines, and 4D reconstruction strategies toward a more powerful geometry foundation model with improved long-horizon scalability, global geometric consistency, and dynamic-scene understanding. We also expect \textsc{UniT} to benefit a wide range of downstream applications, such as navigation \cite{sariyildiz2026kinaema} and robotic policy learning\cite{han2026geometric}.}

\section*{Statements and Declarations}
\subsection*{Funding}

This work was supported by the National Key Research and Development Program of China under Grant 2024YFB4707603, the National Natural Science Foundation of China under Grants U24A20252 and 62373298, and the Guangdong Provincial Key Lab of Integrated Communication, Sensing and Computation for Ubiquitous Internet of Things under Grant 2023B1212010007.

\subsection*{Competing interests}
The authors have no relevant financial or non-financial interests to disclose.

\subsection*{Data availability}
The publicly available datasets used for training and evaluation are identified and cited in \cref{sec:exp}, including the training datasets in \cref{tab:traindata}. Access to these datasets is provided by their respective owners under the applicable terms of use.

\subsection*{Code availability}
The source code for \textsc{UniT} is publicly available through the project page: \url{https://omega-hkustgz.github.io/projects/unit}.

\bibliography{main}

\end{document}